\documentclass{article} % For LaTeX2e
\usepackage{collas2024_conference,times}
\usepackage{easyReview}

\usepackage{comment}
\usepackage{algorithm}
\usepackage{pifont}
\usepackage{algpseudocode}
\usepackage{amsmath}
\usepackage{amssymb}
\usepackage{xcolor}
\usepackage{enumitem}
\usepackage{subcaption}
\usepackage{multirow}
\usepackage[pagebackref=true,breaklinks=true,letterpaper=true,colorlinks,bookmarks=false]{hyperref}
\usepackage{wrapfig}

\newcommand{\Break}{\State \textbf{break} }

\usepackage{amsmath,amsfonts,bm}

\def\eqref#1{equation~\ref{#1}}
\def\1{\bm{1}}

\DeclareMathAlphabet{\mathsfit}{\encodingdefault}{\sfdefault}{m}{sl}
\SetMathAlphabet{\mathsfit}{bold}{\encodingdefault}{\sfdefault}{bx}{n}

\usepackage{hyperref}
\hypersetup{
    colorlinks=true,
    linkcolor=red,
    filecolor=magenta,
    urlcolor=blue,
    citecolor=purple,
    pdftitle={Overleaf Example},
    pdfpagemode=FullScreen,
    }

\title{GRASP: A Rehearsal Policy for Efficient Online Continual Learning}

\author{Md Yousuf Harun\\
Rochester Institute of Technology\\
United States of America\\
\texttt{mh1023@rit.edu} \\
\And % Use And to have authors side by side
Jhair Gallardo  \\
Rochester Institute of Technology\\
United States of America\\
\texttt{gg4099@rit.edu} \\
\And % Use AND to have authors block one under the other
Junyu Chen \\
University of Rochester\\
United States of America\\
\texttt{jchen175@ur.rochester.edu} \\ 
\And % Use AND to have authors block one under the other
Christopher Kanan \\
University of Rochester \\
United States of America\\
\texttt{ckanan@cs.rochester.edu}\\
}

\preprintcopy % Uncomment for the preprint version, but NOT for submission.

\begin{document}

\maketitle

\begin{abstract}
Continual learning (CL) in deep neural networks (DNNs) involves incrementally accumulating knowledge in a DNN from a growing data stream. A major challenge in CL is that non-stationary data streams cause catastrophic forgetting of previously learned abilities. A popular solution is rehearsal: storing past observations in a buffer and then sampling the buffer to update the DNN. Uniform sampling in a class-balanced manner is highly effective, and better sample selection policies have been elusive. Here, we propose a new sample selection policy called GRASP that selects the most prototypical (easy) samples first and then gradually selects less prototypical (harder) examples. GRASP has little additional compute or memory overhead compared to uniform selection, enabling it to scale to large datasets. Compared to 17 other rehearsal policies, GRASP achieves higher accuracy in CL experiments on ImageNet. Compared to uniform balanced sampling, GRASP achieves the same performance with 40\% fewer updates. We also show that GRASP is effective for CL on five text classification datasets. Source code for GRASP is available at \url{https://yousuf907.github.io/graspsite}.
\end{abstract}

\section{Introduction}
\label{intro}

In deep continual learning (CL), a deep neural network (DNN) is sequentially updated from a growing data stream, where the distribution of the data stream is unknown. When the stream is non-stationary, catastrophic forgetting of previously learned abilities occurs if the DNN is progressively fine-tuned with only new samples. CL methods aim to overcome this obstacle. One of the best CL methods is rehearsal (e.g., experience replay)~\citep{van2022three,zhou2023deep}. Rehearsal is highly effective across CL scenarios, and it is especially effective for class incremental learning (CIL), where classes must be learned sequentially~\citep {rebuffi2017icarl}. Rehearsal methods store a subset of old examples, or representations of those examples, in a buffer and then update the DNN with a chosen mixture of new and old data. A \emph{rehearsal policy} governs which samples are selected. The most common approach is uniform sampling, but better policies can potentially reduce the time required for rehearsal. While some have been shown to reduce the total number of updates needed on small-scale CL problems~\citep{aljundi2019online,aljundi2019gradient,bang2021rainbow,yoononline,koh2021online,shim2021online}, sampling is expensive, resulting in no improvement in the total amount of time required for training. One could simply use more updates with uniform selection. 

For large-scale problems, uniform selection has been shown to outperform more sophisticated policies~\citep{harun2023siesta, prabhu2023computationally}. It is also a highly effective strategy for active learning and dataset pruning~\citep{sorscher2022beyond, evans2023bad}.  
Uniform selection is particularly effective in CL settings where the buffer is highly constrained, unlike compute.
This is because, over a large number of training steps, uniform selection will eventually achieve optimal accuracy. In such cases, an ideal selection policy may not provide added value. 
However, when compute is limited and the buffer is large, only a subset of examples can be chosen due to computational constraints. In this scenario, a sample selection strategy plays a crucial role in achieving optimal accuracy over a small number of training steps. 
In this paper, we aim to identify a computationally efficient rehearsal policy that maximizes DNN accuracy with a fixed number of updates and a relatively large buffer. Our setting also aligns with the industry, where computational costs significantly exceed storage costs~\citep{prabhu2023computationally}.

%Despite these findings, we argue that policies that enable more efficient learning should exist. 
Our method, GRASP, is inspired by dataset pruning, where the goal is to identify a sample subset that when trained on has the same performance as the entire dataset. \citet{sorscher2022beyond} showed that the optimal pruning strategy varied based on the size of the dataset. Based on similarity to class prototypes, they found retaining the easiest samples was the best strategy for small datasets, whereas for large datasets keeping the hardest samples was best. These observations were made based on having a fixed dataset, where the size of the dataset is known beforehand. 
%\textcolor{magenta}{In contrast, in CL, the size of the dataset and the number of samples for each class change over time.} 
%\replace{In contrast, in CL, the size of the dataset and the number of samples for each class increase over time.} {In contrast, in CL, the size of the dataset and the number of samples for each class change over time.}
In contrast, in CL, the size of the dataset and the number of samples for each class change over time.
This suggests that an effective rehearsal policy should adapt to the amount of currently available data for a class.

\paragraph{This paper makes the following key contributions:}
\begin{enumerate}[noitemsep, nolistsep]
    \item We propose GRASP, a dynamic rehearsal policy that initially selects easy samples and gradually selects harder ones. GRASP is compute and memory efficient. We integrated GRASP into three rehearsal-based CL systems, SIESTA~\citep{harun2023siesta}, DERpp~\citep{buzzega2020dark}, and GDumb~\citep{prabhu2020gdumb}.
    %\item In CIL on ImageNet, GRASP outperforms 17 other rehearsal policies, including uniform balanced.
    \item In CIL experiments on ImageNet, GRASP outperforms 17 other rehearsal policies, including class-balanced uniform selection.
    \item We show that GRASP is effective across CL distributions, including independent and identically distributed (IID) and long-tailed distributions.    
    \item We demonstrate that GRASP is effective for natural language processing (NLP) on 5 CL benchmarks.
\end{enumerate}

\begin{figure}[t]
  \centering

\begin{subfigure}[b]{0.4\textwidth}
         \centering
         \includegraphics[width=\textwidth]{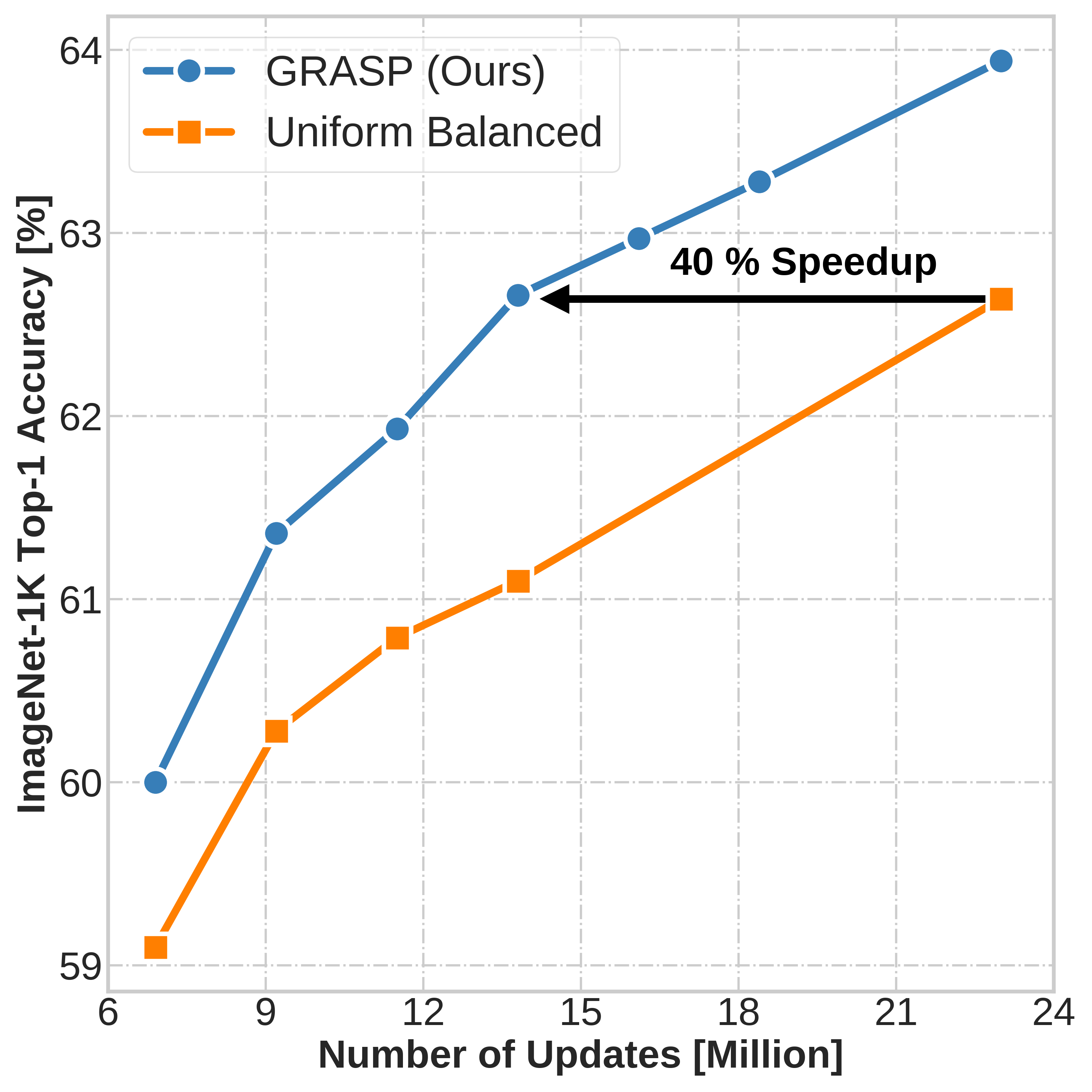}
         \caption{Compute Comparison}
         \label{fig:compute}
     \end{subfigure}
     \hfill
      \begin{subfigure}[b]{0.4\textwidth}
         \centering
         \includegraphics[width=\textwidth]{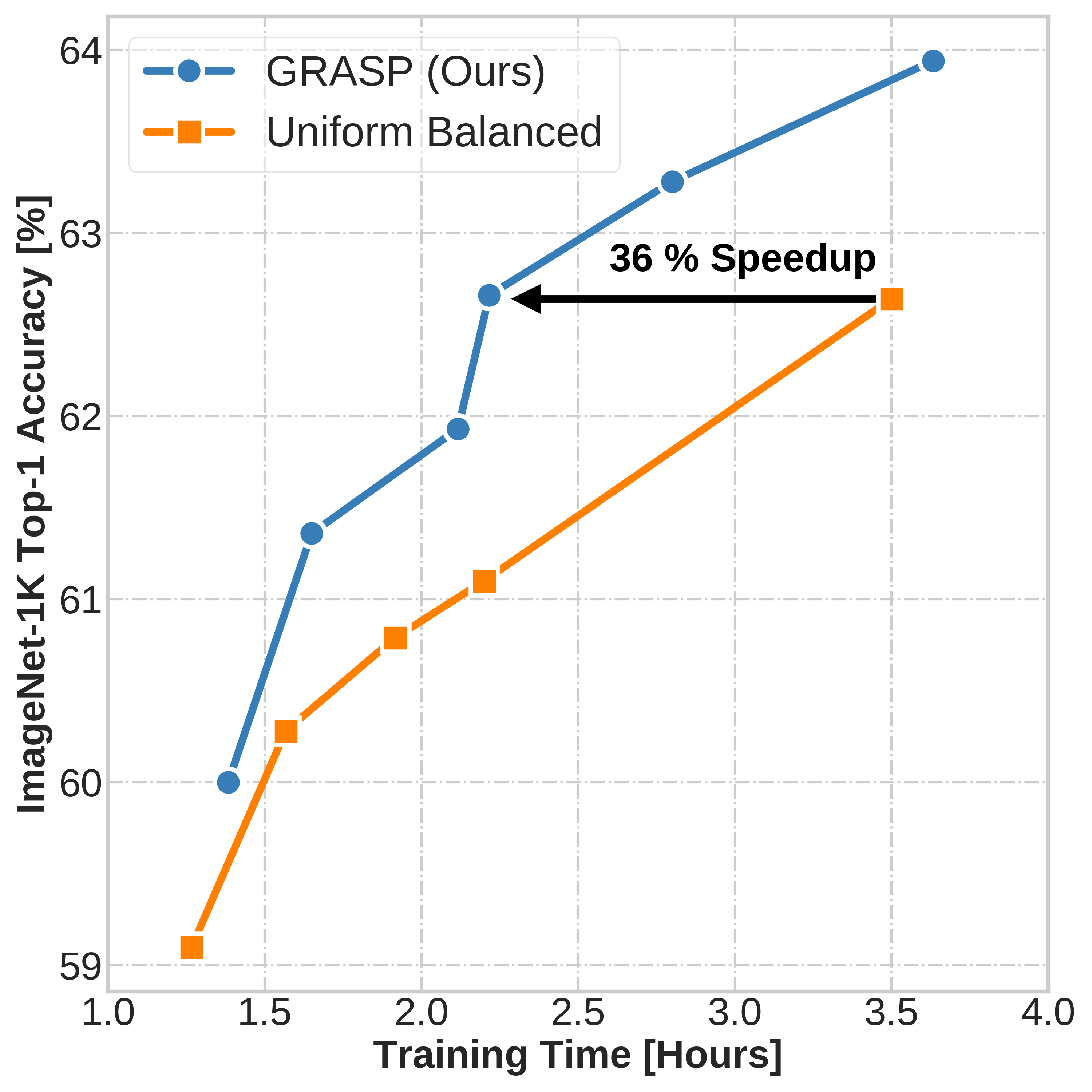}
         \caption{Time Comparison}
         \label{fig:train_time}
     \end{subfigure}
   \caption{%The GRASP rehearsal policy outperforms 17 others in CIL. \jg{this phrase doesn't make sense with the plots here}
   GRASP achieves the best accuracy of the popular uniform balanced policy while requiring $40\%$ fewer gradient descent updates and $36\%$ less training time for CIL with SIESTA on ImageNet-1K. %\jg{I think more details here are needed. What method (SIESTA?). What task? CIL? Alos, maybe we can mention this figure in the contributions somehow} \yh{done}     
   }
   \label{fig:compute_time}
   \vspace{-1em}
\end{figure}

%\newpage 
\section{Related Work}

\subsection{Rehearsal-Based Continual Learning}

Rehearsal is inspired by neuroscience, where recent memories are stored in the hippocampus and then reactivated for long-term consolidation~\citep{o2010play, hayes2021replay}. Most CL methods use a rehearsal buffer where the maximum size is constrained~\citep{hayes2021replay, chaudhryER_2019, abraham2005memory, belouadah2019il2m, castro2018end, chaudhry2018efficient, chaudhry2018riemannian, hayes2019memory, hayes2020REMIND, harun2023siesta, rebuffi2017icarl, tao2020topology, wu2019large, aljundi2019gradient}. These constraints are often arbitrary and require having a policy for maintaining the size constraint (see Sec.~\ref{sec:buffer-policy}). 
However, some early CL methods used \emph{cumulative rehearsal} where rehearsal buffers store all previously observed samples~\citep{gepperth2016bio,hayes2019memory,kemker2018measuring}. Recently, there has been a resurgence of interest in this setting because it allows one to focus on how to utilize the buffer best, e.g., to minimize training time while maximizing accuracy~\citep{al2023rapid,prabhu2023computationally,prabhu2023online,harun2023overcoming, verwimp2024continual}. We study both settings in this paper.

Rehearsal methods fall into three categories: veridical, latent, and generative. In \textbf{veridical rehearsal}, raw input data is stored in a memory buffer for later rehearsal~\citep{rebuffi2017icarl, chaudhryER_2019, castro2018end, lopez2017gradient, bang2021rainbow, chaudhry2018riemannian, wu2019large, aljundi2019gradient}. Instead of storing raw images, \textbf{latent rehearsal} methods store features from hidden layers~\citep{hayes2020REMIND, iscen2020memory, caccia2020online, pellegrini2020latent, zhao2021memory, harun2023siesta}, allowing them to store far more samples than veridical under the same memory budget. If storing data is prohibited, \textbf{generative rehearsal} methods produce synthetic images or features to retrieve old knowledge~\citep{shin2017continual, he2018exemplar, hu2018overcoming, liu2020generative, kemker2017fearnet, ostapenko2019learning, xiang2019incremental} by incorporating a generator into the DNN; however, these methods increase compute and often struggle with feature drift. While most work on rehearsal has focused on images, these three types have also been used for CL in NLP~\citep{ke2022continual, biesialska2020continual}. In our work, we conduct experiments in both the veridical and latent rehearsal settings.

\subsection{Rehearsal \& Buffer Maintenance Policies}
\label{sec:buffer-policy}

\begin{table}[t]
  \caption{
   \textbf{Advanced Rehearsal Policies.} Current state-of-the-art methods except MIR and ASER mainly proposed buffer maintenance policy and used uniform random as rehearsal policy. %(assuming all samples in the buffer are equally informative for rehearsal). 
   %Unlike GRASP, these methods are computationally expensive and difficult to scale. %%Most methods including MIR and ASER partially use uniform random to reduce computational overhead which questions the standalone contribution of the proposed policies.
   }
  \centering
  \resizebox{\linewidth}{!}{
     \begin{tabular}{cccccc}
     \hline
     \textbf{Method} & \textbf{Buffer Policy} & \textbf{Rehearsal Policy} & \textbf{Metric} & \textbf{Expensive} & \textbf{Scalable} \\
     \hline
     MIR~\citep{aljundi2019online} & \ding{55} & \ding{52} & MIR Loss & \ding{52} & \ding{55} \\
     Rainbow Memory~\citep{bang2021rainbow} & \ding{52} & \ding{55} & Uncertainty & \ding{52} & \ding{55} \\
     OCS~\citep{yoononline} & \ding{52} & \ding{55} & OCS & \ding{52} & \ding{55} \\
     GSS~\citep{aljundi2019gradient} & \ding{52} & \ding{55} & Grad Variance & \ding{52} & \ding{55} \\
     Grad Matching~\citep{campbell2019automated} & \ding{52} & \ding{55} & Gradient & \ding{52} & \ding{55} \\
     Bi-level~\citep{borsos2020coresets} 
     & \ding{52} & \ding{55} & Bi-level Opt & \ding{52} & \ding{55} \\
     CLIB~\citep{koh2021online} & \ding{52} & \ding{55} & Max Loss & \ding{52} & \ding{55} \\
     ASER~\citep{shim2021online} & \ding{52} & \ding{52} & Shapley Value & \ding{52} & \ding{55} \\
     %RMM~\cite{liu2021rmm} & \ding{52} & \ding{55} & RMM & \ding{52} & \ding{55} \\
     %InfoRS~\cite{sun2021information} & \ding{52} & \ding{55} & InfoRS & \ding{52} & \ding{55} \\
     \hline
     \textbf{GRASP} (Ours) & \ding{55} & \ding{52} & Cosine Distance & \ding{55} & \ding{52} \\
     \hline
    \end{tabular}}
  \label{tab:policy_type}
\end{table}

Rehearsal algorithms alternate between a sample acquisition phase, where samples are added to a buffer, and a rehearsal phase, where the DNN is updated using the buffer and newly acquired samples. All rehearsal algorithms must then define a \emph{Rehearsal Policy} for what should be sampled from the buffer. For memory-constrained rehearsal, this is often entangled with what we call the \emph{Buffer Maintenance Policy}, which defines what should be kept within the buffer. In many memory-constrained methods, the entire buffer is used to update the network, where the rehearsal policy is then implicitly governed by the criteria used to determine what is kept within the buffer. This work aims to disentangle these two aspects by converting buffer maintenance policies into rehearsal policies to compare approaches. We summarize the properties of recent methods in Table~\ref{tab:policy_type}. %Unlike GRASP, the listed methods are computationally expensive and difficult to scale.

%\paragraph{Rehearsal Policies.} 
\textbf{Rehearsal Policies.} 
While buffer maintenance policies have been heavily explored in CIL, much less work has been done to identify which stored samples should be selected for rehearsal (see Table~\ref{tab:policy_type}). We briefly describe the rehearsal policies that have been studied. %\ck{What about all the methods from Tyler's paper we study? They aren't mentioned here. It seems like a sentence is needed to summarize that many simple policies have been studied, which we describe later, but that we will go over two more sophisticated ones.} 
Prior CL work~\citep{hayes2021selective,chaudhry2018riemannian,harun2023siesta,prabhu2023computationally} studied a variety of sampling policies e.g., uniform balanced, min rehearsal, max loss, min margin, min logit-distance, and min confidence; however, these policies showed limited efficacy for large-scale datasets. More details about these policies are given in Sec~\ref{subsec:compared_rehearsal_policies}.
More advanced rehearsal policies for CL include MIR and ASER.
In MIR, when a new batch of data arrives, virtual updates are made to the DNN using the new batch to find the maximally interfered old samples~\citep{aljundi2019online}. Next, it updates DNN using the new batch mixed with interfered old data. Virtual updates are computationally expensive, and MIR disproportionately prioritizes redundant samples in the most interfered category~\citep{shim2021online}. ASER uses Shapley values to prioritize representative samples for storage and interfered samples for rehearsal~\citep{shim2021online}. ASER uses uniform random sampling to construct evaluation and candidate sets to reduce computational overhead, and it is difficult to scale since it computes the Euclidean distance of each candidate sample from each evaluation sample at every training iteration. Recently,~\citet{prabhu2023computationally} compared many rehearsal policies and found that balanced uniform outperforms others for large-scale datasets.

\textbf{Buffer Maintenance Policies.} The most commonly used method is reservoir sampling, where a new sample overwrites a randomly selected sample from the buffer once the buffer is full, and then samples are chosen uniformly from the buffer for rehearsal. This strategy is used in both vision~\citep{wang2023comprehensive,riemer2019learning, chaudhry2019tiny} and NLP~\citep{ke2022continual, biesialska2020continual}. More advanced strategies exploit the statistics of the stored samples e.g., GDumb~\citep{prabhu2020gdumb}, ExStream~\citep{hayes2019memory}, ring buffer~\citep{lopez2017gradient}, herding-based~\citep{rebuffi2017icarl}, $k$-Means~\citep{chaudhry2019tiny}, MoF~\citep{chaudhryER_2019}, and FIFO~\citep{lopez2017gradient}.
An alternative to data-driven methods are model-based methods that determine what to retain in the buffer by analyzing the DNN's behavior on the stored samples~\citep{chaudhry2018riemannian,koh2021online,borsos2020coresets,campbell2019automated,aljundi2019gradient,yoononline}. For example, rainbow memory~\citep{bang2021rainbow} stores diverse samples based on classification uncertainty and image augmentation.
However, model-based methods, especially gradient-based methods, are computationally expensive and intractable for large-scale datasets e.g., ImageNet-1K. 
For instance, OCS~\citep{yoononline} is computationally demanding since at every training iteration it scores data based on mini-batch gradient similarity and cross-batch gradient diversity.
See \citet{wang2023comprehensive} for a review. We repurposed buffer maintenance policies for rehearsal policies to compare them with GRASP. More details are given in Sec~\ref{subsec:compared_rehearsal_policies}.

%\input{tables/policy_type}

% \clearpage

%%%%%%%%%%%%%%%%%%%%%%%%%%%%%%%%%%%%%%%%%%%%
\section{The GRASP Rehearsal Policy}
\label{sec:grasp_rehearsal_policy}
%% yh{Algorithm can be wrapped around the text for space.}

% \centering
%\begin{wrapfigure}[25]{t}{0.5\textwidth}%{r}{9cm}
\begin{wrapfigure}[26]{t}{0.5\textwidth}%{r}{9cm}
%\begin{wrapfigure}[25]{t}{0.5\textwidth}%{r}{9cm}
\vspace{-0.65cm}
\begin{minipage}{\linewidth} %.65\linewidth

    \begin{algorithm}[H]
    
    \caption{\textbf{GRASP Rehearsal Policy}}
    \label{alg:grasp}
    
        \begin{algorithmic}
            \Require $\mathcal{M} = \{(\mathcal{X}, \mathcal{D})_{k}\}_{k=1}^{K}$
            \Comment{Memory buffer}
            \State $\mathcal{R} \gets \{\}$ \Comment{Will contain $\mathcal{U}$ selected samples}
            \State $c \gets 0$
            \Comment{Initialize counter}
            \State $\mathcal{U} = n \times b$ \Comment{Compute Budget}
            
            \While{$c < \mathcal{U}$}
                \For{$k$ = 1 to $K$}
                    \State $\mathcal{X}_{k}, \mathcal{D}_{k} \gets \mathcal{M}$ \Comment{Obtain data for class $k$}
                    \State $P_{k} = \mbox{normalize} \left( \mathcal{D}_{k}^{-1} \right)$ \Comment{Compute probabilities}
                    \State $m \overset{1}{\sim} \mbox{sample} (\mathcal{X}_{k}, P_{k})$ 
                    \State $\mathcal{D}_{k}[m] \gets \mathcal{D}_{k}[m]+\mbox{max}(\mathcal{D}_{k})$ \Comment{Virtual update}
                    %%%%%%%% original %%%%%%%%%%
                    % \State $\{\mathbf{x}, k\} \gets \mathcal{X}_{k}[m]$ \Comment{Obtain sample}
                    % \State $\mathcal{R} \gets \{\mathbf{x}, k\}$ \Comment{Add sample}
                    %%%%%%%%%%%%%%%%%%%%%
                    
                    \State $(\mathbf{x}, k) \gets \mathcal{X}_{k}[m]$ \Comment{Obtain sample}
                    \State $\mathcal{R} \gets \mathcal{R} \cup \{(\mathbf{x}, k)\}$ \Comment{Add sample}
                    
                    \State $c \gets c+1$ \Comment{Update counter}
                    
                    \If{$c \geq \mathcal{U}$}
                        \Break
                    \EndIf
                    
                \EndFor
            \EndWhile
            
            \For{$t$ = 1 to $b$}
                %%%%%%%% original %%%%%%%%%%
                % \State $B_{t} \overset{n}\sim \mathcal{R}$ \Comment{Select a mini-batch of size n}
                %%%%%%%%%%%%%%%%%%%%%%%%

                \State $B_{t} \gets \mathcal{R}[(t-1) n: t n]$ \Comment{Select a mini-batch of size n}

                \State $\theta \gets \mbox{SGD}(\theta, B_{t})$ \Comment{Update model on $n$ samples}
            \EndFor \Comment{Rehearsal cycle ends}
            %\State Reset and recompute \mathcal{D}_{k}
        \end{algorithmic}
    \end{algorithm}

\end{minipage}
\end{wrapfigure}
% \par

We propose a new rehearsal policy named GRASP (\textbf{\textcolor{orange}{GRA}}dually \textbf{\textcolor{orange}{S}}elect less \textbf{\textcolor{orange}{P}}rototypical). 
GRASP is simple, scalable, hyperparameter-free, and has little additional compute or memory overhead compared to
uniform selection. Our goal is to identify a rehearsal policy that minimizes the number of DNN updates to gain computational efficiency.

GRASP is based on the hypothesis that choosing only easy or hard samples are both suboptimal and that the DNN would benefit from a curriculum that combines both. GRASP first selects the most prototypical (easy) samples from the buffer and then gradually selects harder samples, where easy samples are closest to the class mean and hard samples are farthest. %(hard) samples are nearest to (furthest from) class-mean and 2) avoids rehearsing interfered samples early in training since samples near class mean interfere less with samples from different classes.
While prior work has explored policies that select samples close to prototypes for buffer maintenance~\citep{rebuffi2017icarl,chaudhryER_2019}, their policies are biased toward easy samples near class prototypes instead of progressively choosing hard samples as well.

GRASP can be integrated into both online and offline rehearsal-based CL methods, and it can be used with either veridical or latent rehearsal. In all cases, we assume that the CL algorithm maintains a buffer containing both old and new samples. In offline CL memory constraints on the buffer only apply to old samples, where all new samples are added to it before rehearsal, and then after rehearsal, the buffer is compressed to the given memory budget. Severe catastrophic forgetting in minority classes can occur in rehearsal due to class-imbalance~\citep{wu2019large}, which GRASP overcomes by selecting samples in a class-balanced way, where minority classes can be oversampled.

%%%%%%%%%%%%%%%%%%%%%%%%%
\paragraph{Sample Acquisition Phase.}
%\textbf{Sample Acquisition Phase.}
For the CL algorithms we study, a sequence of labeled samples are provided to the learner over time, where after a sufficiently large number of samples arrive the learner uses rehearsal to update the DNN. To simplify notation, we assume at time $t$ the learner receives a single sample $X_t$ with label $k_t$, where $X_t$ is the raw input in veridical rehearsal or an embedding from the middle of the network in latent rehearsal. The sample is then added to the buffer, where a sample from the largest class is removed from the buffer if it is full.

After buffer maintenance, we compute the distances between stored samples and class prototypes for GRASP.
Let $\mathbf{z} \in \mathbb{R}^d$ be the embedding computed by the DNN from the penultimate layer.
For each class $k$, a class prototype vector $\mathbf{q}_{k}$ is computed by averaging the penultimate embedding vectors of corresponding samples $\mathcal{X}_k$, i.e., $\mathbf{q}_{k} = \frac{1}{J} \sum_{j=1}^{J} \mathbf{z}_{k, j}$. Next for each sample, the cosine distance $d$ between its penultimate embedding and the class prototype is calculated as $d = 1 - (\mathbf{z} \cdot \mathbf{q}_{k}) / ({\left\| \mathbf{z}\right\| _{2}\left\| \mathbf{q}_{k}\right\| _{2})}$.
%computed
% \begin{equation}
%     d = 1 - \dfrac {\mathbf{z} \cdot \mathbf{q}_{k}} {\left\| \mathbf{z}\right\| _{2}\left\| \mathbf{q}_{k}\right\| _{2}}.
% \end{equation}
%The distance is used during rehearsal to select samples based on how similar they are to the prototype and is stored in the buffer. 
The distance $d$ is used during rehearsal to select samples based on how far they are from the prototype and is stored in the buffer.
Therefore, the buffer consists of $\mathcal{M} = \{(\mathcal{X}, \mathcal{D})_{k} \}_{k=1}^{K}$, where $K$ is the total number of classes that have been seen, $\mathcal{X}_k$ is the set of stored inputs from class $k$ for veridical rehearsal or stored embeddings for latent rehearsal, and $\mathcal{D}_k$ has the cached distances to the class prototypes.

%%%%%%%%%%%%%%%%%%%%%%%%%
\paragraph{Rehearsal Phase.}
%\textbf{Rehearsal Phase.}
Because the focus of this work is computational efficiency, the rehearsal phase is given a compute budget of $\mathcal{U}=nb$ gradient updates to the DNN, where $b$ is the total number of minibatches and $n$ is the minibatch size.  GRASP iteratively selects $\mathcal{U}$ samples from the buffer. For class $k$, GRASP assigns a selection probability $P_k$ inversely proportional to $\mathcal{D}_{k}$ for all the data points in class $k$. The selected sample's distance is then temporarily updated by adding $\max \mathcal{D}_{k}$ to it. This makes it unlikely for that sample to be chosen again in the rehearsal session or task because it will have the lowest probability; therefore, within a rehearsal session, GRASP chooses progressively less prototypical samples for rehearsal. Approximately the same number of samples are chosen from each class, where oversampling can occur for classes with fewer samples. After the rehearsal session ends and a new session begins, the distances among samples and class prototypes, $\mathcal{D}_{k}$ are recomputed. Pseudocode for GRASP is given in Algorithm~\ref{alg:grasp}.

%%%%%%%%%%%%%%%%%%%%%%%%%%%%%%%%%%%%%%%%%%
\section{Experimental Setup}
\label{sec:exp_setup}

In experiments, we study rehearsal policies in both the unbounded memory and the conventional memory-bounded settings. The unbounded setting enables us to focus on the efficacy of rehearsal policies regardless of the amount of storage permitted or the choice of buffer maintenance policy. Moreover, for industrial applications, the cost of deep learning largely depends on computing rather than storage. 
The unlimited memory setting has been studied in recent CL papers that have argued it is better aligned with industry needs~\citep{Harun_2023_CVPR, harun2023overcoming,  prabhu2023computationally, prabhu2023online, bornschein2022nevis, al2023rapid, verwimp2024continual}.

\subsection{CL Datasets and DNN Architectures}% Orderings}
To show scalability to large-scale datasets, our main results use \textbf{ImageNet}~\citep{russakovsky2015imagenet}. %, where the DNN takes as input a $224 \times 224$ image. 
%It has 1000 categories, each with $732-1300$ training images and $50$ for validation. 
We conduct CIL experiments (Non-IID ordering) with the 1000 class version of the dataset (ImageNet-1K) and two variants with 150 and 300 classes, referred to as ImageNet-150 and ImageNet-300 respectively. Long-tailed datasets are challenging in CIL, and to assess rehearsal policies in this setting we use \textbf{Places-LT}~\citep{liu2019large}. 
Additionally, we study IID orderings on ImageNet-1K to assess the robustness of rehearsal policies to non-adversarial data streams. More dataset details are included in Appendix~\ref{sec:dataset}.
%\yh{We can probably shorten this paragraph and move dataset details to Appendix for space.}
Continual text classification performs task incremental learning. Following \citet{huang2021continual}, we use five large-scale benchmark text datasets: \textbf{AG News} (news classification), \textbf{Yelp} (sentiment analysis), \textbf{DBPedia} (Wikipedia article classification), \textbf{Amazon} (sentiment analysis), and \textbf{Yahoo! Answer} (Q$\&$A classification). See \citet{huang2021continual} for details. Methods are evaluated on 6 task sequences. % (3 length-3 and 3 length-5).

% \subsection{DNN Architectures}

For our main image classification experiments, we use \textbf{MobileNetV3-Large}~\citep{howard2019searching} since it outperforms widely used ResNet-18~\citep{harun2023siesta} and has lower latency. Following~\cite{harun2023siesta}, we pre-trained MobileNetV3-Large on the first 100 classes from ImageNet-1K using SwAV~\citep{caron2020unsupervised} for all experiments. %The pretraining classes do not overlap with ImageNet-300 and ImageNet-150.
In Appendix~\ref{sec:vit}, we also conduct experiments on a vision transformer \textbf{MobileViT-Small}~\citep{mehta2021mobilevit}. We use \textbf{BERT}~\citep{kenton2019bert} in continual text classification. Implementation details are in Appendix~\ref{sec:implement}.

\subsection{Rehearsal-Based CL Algorithms}

We study the %effectiveness of 
rehearsal policies in three algorithms for large-scale CL in vision and one for CL in NLP:
\begin{enumerate}[leftmargin=*]
    \item \textbf{SIESTA} is a state-of-the-art latent rehearsal CL algorithm that alternates between online and offline phases~\citep{harun2023siesta}. SIESTA is initialized on the first 100 classes of ImageNet-1K using self-supervised learning~\citep{gallardo2021self}. During its online phase, it only updates its output layer and stores quantized tensor embeddings of the input images. During its offline phase (every 100 ImageNet classes), it uses latent rehearsal to update its non-frozen layers with a fixed number of updates. %It uses product quantization to store much larger amounts of data than veridical rehearsal methods under the same memory budget. 
    %During each offline phase, it performs a fixed number of updates to the non-frozen layers using rehearsal. 
    We also do experiments with a variant that uses veridical rehearsal.
    We use SIESTA for our main experiments because it requires only 2 hours to finish %end-to-end 
    training on ImageNet-1K on a single GPU and because it matches an offline model's accuracy on ImageNet-1K in the augmentation-free setup. %\jg{I reduced some text}
    
    \item \textbf{Dark Experience Replay (DERpp)} combines rehearsal with knowledge distillation and regularization~\citep{buzzega2020dark}. It uses a distillation loss on logits of old samples for consistency. It uses reservoir sampling to maintain a constrained memory buffer.
    \item \textbf{GDumb} uniformly removes a sample from the largest class upon arrival of a new sample when memory buffer is full~\citep{prabhu2020gdumb}.
    \item \textbf{IDBR} 
    is a CL system for text classification~\citep{huang2021continual}. Using rehearsal, it continually updates a BERT text encoder and a linear classification layer. In the original IDBR system, it also updated regularization sub-networks for retaining task-generic information and for adapting to task-specific information for regularization. We exclude these sub-networks to focus on rehearsal.
\end{enumerate}

In our experiments, all methods use a fixed number of rehearsal updates. Experiments omit image augmentation to focus on comparing rehearsal policies. 

% \begin{wrapfigure}[11]{t}{0.5\textwidth}
%\begin{wrapfigure}[14]{t}{0.5\textwidth}
\begin{wrapfigure}[12]{t}{0.5\textwidth}
    % \begin{figure}[t]
        \vspace{-0.6cm}
        \centering
        \includegraphics[width=\linewidth]{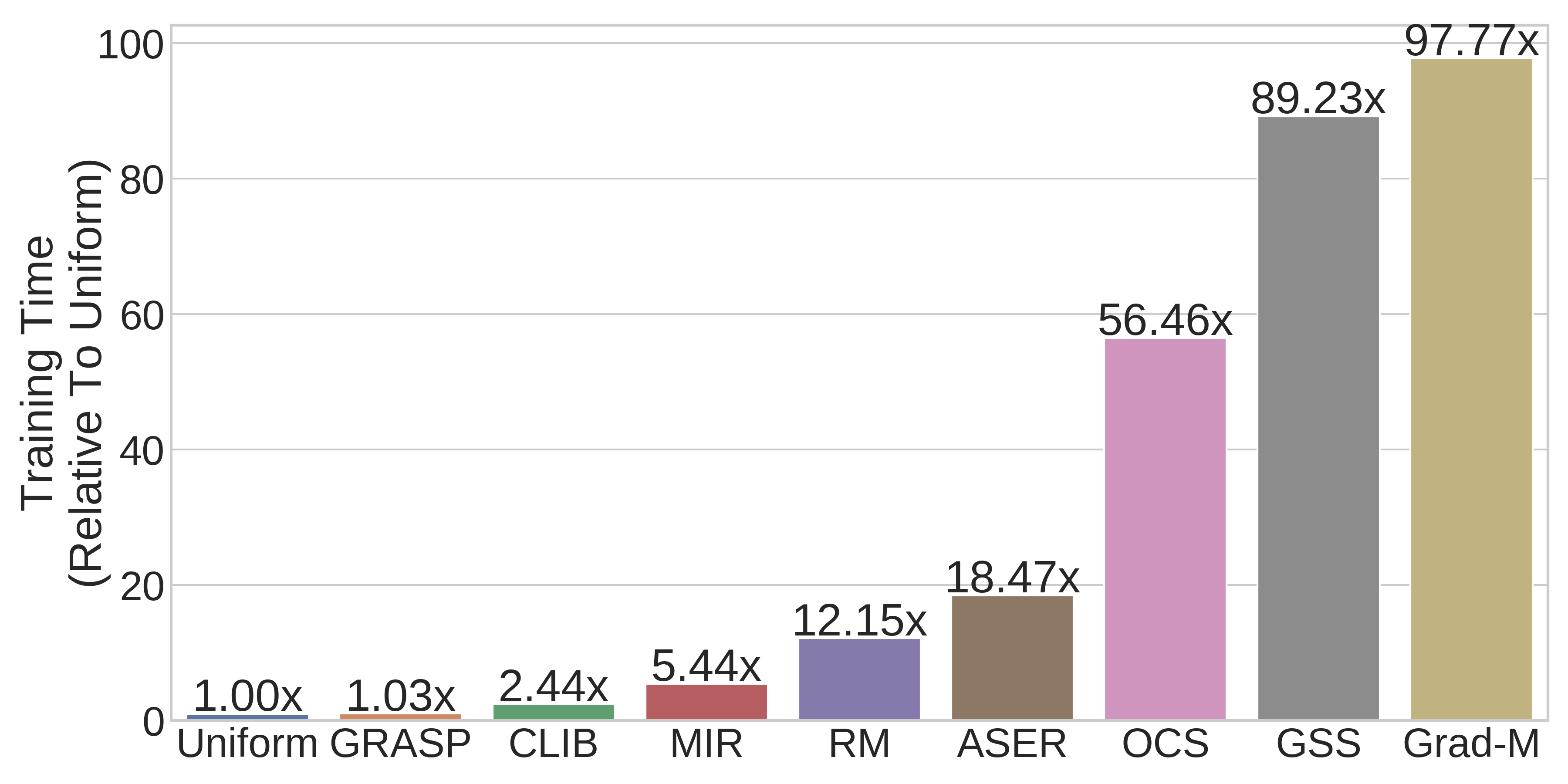}
        \caption{Unlike GRASP, existing state-of-the-art methods require significantly longer training time than uniform.
        }
        \label{fig:relative_train_time}
    % \end{figure}
\end{wrapfigure}

Our main results use SIESTA \citep{harun2023siesta}. SIESTA is pre-trained on the first 100 ImageNet classes. Subsequently, it learns 50/ 200/ 900 additional classes continually. %SIESTA does lightweight updates to its output layer during its online phase. Its offline rehearsal phase occurs every 100 classes or after an equivalent number of samples in our IID experiments.  
For SIESTA with MobileNetV3-Large, the first 8 layers are frozen, and the remaining layers (97.81\% of parameters) are updated during latent rehearsal.
For SIESTA, DERpp, and GDumb, we use versions with latent and veridical rehearsal.  We created latent rehearsal versions of DERpp and GDumb using SIESTA's setup. 

\textbf{Memory Constraints.} We study CL with both unbounded memory (access to the entire dataset) and bounded memory settings. Following~\citet{hayes2020REMIND}, all methods are limited to $1.5$ GB of storage for ImageNet-1K. This corresponds to 10000 old images for veridical rehearsal, which excludes the newly acquired batch of 120000 images. For ImageNet-150/300 subset,  latent rehearsal methods use up to $0.2$ GB for the buffer. For IDBR, we randomly select $50\%$ of seen examples to store in the memory buffer. Samples are removed from the largest class/task to maintain a bounded rehearsal buffer.

\textbf{Compute Constraints.} 
Rehearsal policies use a fixed computational budget ($\mathcal{U}=nb$) that indicates the total number of samples used for backpropagation during rehearsal.
Following SIESTA~\citep{harun2023siesta}, in our main results on ImageNet-300 we set the number of iterations per rehearsal session to $1251$ ($n$) with a mini-batch size $512$ ($b$) for all methods. 
In ImageNet-150 experiments, we use $500$ iterations per rehearsal session with a mini-batch size $64$. 
In ImageNet-1K experiments, we use $2502$ iterations per rehearsal session with a mini-batch size $512$. 
%Details for other datasets and additional experiments with varied compute constraints are included in supplemental material.  
Details for other settings are given in Appendix~\ref{sec:implement}.
For continual text classification with IDBR, we bound compute by using rehearsal sessions of a fixed length, whereas the original IDBR increased the amount of compute as new tasks were learned.

\subsection{Compared Rehearsal Policies and Evaluation Criteria}
\label{subsec:compared_rehearsal_policies}

Here, we briefly describe all the rehearsal policies that we consider as baselines. Note that we convert buffer maintenance policies %e.g., Rainbow Memory, CLIB, GSS, Grad Matching, and OCS 
into rehearsal policies by using them to score stored samples for selection during rehearsal. All baseline methods use the same buffer maintenance policy as described in Sec.~\ref{sec:grasp_rehearsal_policy} where a randomly chosen old sample from the largest class is removed to accommodate a new sample when the buffer is full.

% We compare GRASP with the following policies:
\begin{enumerate}[noitemsep,nolistsep,leftmargin=*]
    \item \textbf{Uniform:} samples are selected uniformly at random~\citep{vitter1985random}.
    \item \textbf{Uniform Balanced:} samples are selected uniformly at random with an equal number of samples per category~\citep{prabhu2020gdumb}. This is a strong baseline for large-scale CL~\citep{prabhu2023computationally}.
    \item \textbf{Min Rehearsal:} samples with least rehearsal count(s) are most likely to be selected~\citep{hayes2021selective}.
    \item \textbf{Max Loss:} samples with higher (lower) cross-entropy loss are defined as hard (easy) samples and prioritized for selection~\citep{kawaguchi2020ordered}.
    \item \textbf{Min Margin:} easy and hard examples are defined with margins, where the margin is the difference between the predicted and correct class probabilities. Hard samples are more likely to be chosen~\citep{scheffer2001active}.
    \item \textbf{Min Logit-Distance:} 
    samples closer to the decision boundary are selected~\citep{chaudhry2018riemannian}.
    \item \textbf{Min Confidence:} samples with lower DNN confidence (softmax scores) are prioritized~\citep{gal2017deep}. 
    \item \textbf{$k$-Means:} features from the penultimate layer are clustered into $k$ centroids with $k$-Means. Samples closer to centroids are more likely to be sampled~\citep{chaudhryER_2019, prabhu2023computationally}. %In GRASP, samples near class mean are most representative.
    \item \textbf{MoF / Easy Biased:} prioritizes samples near the class mean~\citep{chaudhryER_2019}.
    \item \textbf{Hard Biased:} prioritizes samples far from class mean.
    \item \textbf{Rainbow Memory:} %maintains a diverse set of examples per class based on uncertainty and augmentation~\citep{bang2021rainbow}.
    keeps class diverse examples based on uncertainty and augmentation~\citep{bang2021rainbow}.
    \item \textbf{MIR:} %Maximal Interfered Retrieval (MIR)
    prioritizes old samples from %memory 
    buffer that maximally interfere with new samples~\citep{aljundi2019online}.
    \item \textbf{CLIB:} %Continual Learning for i-Blurry (CLIB)
    prioritizes samples with maximum loss decrease~\citep{koh2021online}. 
    \item \textbf{ASER:} %Adversarial Shapley Value Experience Replay (ASER) 
    selects samples based on adversarial Shapley values (SVs)~\citep{shim2021online}. Positive SVs w.r.t. memory samples indicate representative samples whereas negative SVs w.r.t. new samples indicate interfered samples.
    \item \textbf{OCS:} %Online Coreset Selection (OCS) 
    uses mini-batch gradient similarity and cross-batch gradient diversity for sample selection~\citep{yoononline}.
    \item \textbf{GSS:} %Gradient-based Sample Selection (GSS) 
    frames the sample selection as a constraint selection problem to maximize the variance of gradient direction~\citep{aljundi2019gradient}.
    \item \textbf{Grad Matching:} selects samples using Hilbert coreset~\citep{campbell2019automated}.
\end{enumerate}

\paragraph{Evaluation Criteria.}
%\textbf{Evaluation Criteria.}
For evaluation, we use the average accuracy $\mu$ over all rehearsal sessions $T$, where $\mu = \frac{1}{T} \sum_{t=1}^{T} \alpha_{t}$, with $\alpha_{t}$ denoting the accuracy at rehearsal session $t$. 
We use $\mu_{N}$, $\mu_{O}$, and $\mu_{A}$ to denote average accuracy $\mu$ on new, old, and all classes respectively. 
%\textcolor{magenta}{To assess model's ability to balance stability-plasticity, we report $\mu_{N}$ and $\mu_{O}$.}
Final accuracy on all classes is represented by $\alpha$. All metrics use top-1 accuracy ($\%$). 
%We report offline accuracy (upper bound) based on offline models trained on the entire dataset. 
%Rehearsal policies use a fixed computational budget ($\mathcal{U}$) that indicates the total number of samples used for backpropagation during rehearsal.
For continual text classification, following IDBR~\citep{huang2021continual} our evaluation takes place after training models on all tasks and the average across on all test sets is reported.

\section{Results}
\label{results}
We compare GRASP with a variety of 14 rehearsal methods on ImageNet-300 in Sec.~\ref{sec:imagenet_subset_main}. We also compare GRASP with 3 gradient-based methods on ImageNet-150 in Sec.~\ref{sec:grad_main}. Next, we evaluate GRASP on full ImageNet-1K in various settings in Sec.~\ref{sec:imagenet1k_main}. Finally, we summarize additional supporting results in Sec.~\ref{sec:additional_exp}.

%\input{figures/replays} % showing ImageNet-1K curves for GRASP and uniform balanced in both veridical and latent rehearsals

%\subsection{Rehearsal Experiments on ImageNet-300}
\subsection{GRASP vs. Various Rehearsal Methods}
\label{sec:imagenet_subset_main}
\begin{figure}[t]
  \centering

\begin{subfigure}[b]{0.32\textwidth}
         \centering
         \includegraphics[width=\textwidth]{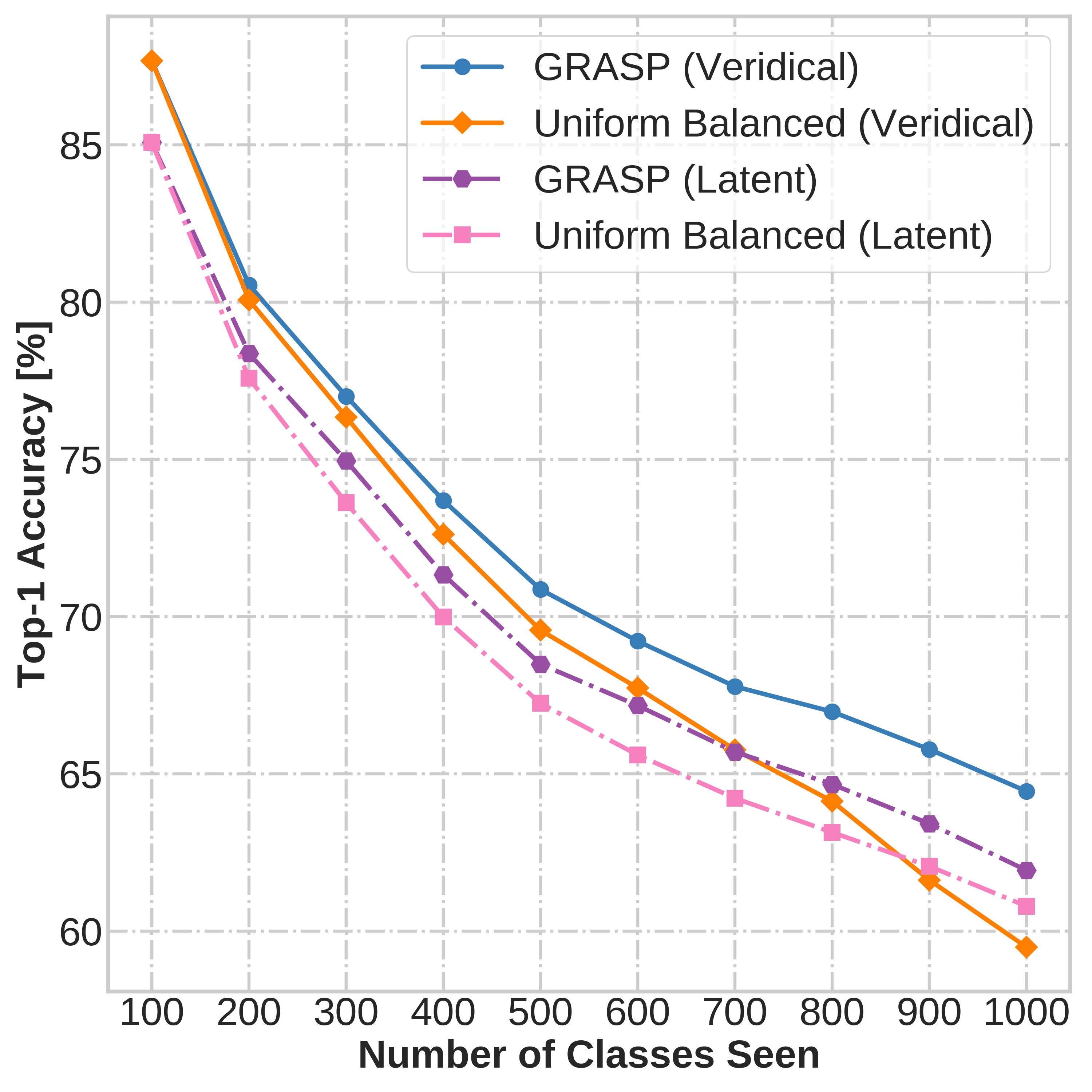}
         \caption{ImageNet-1K Learning Curves}
         \label{fig:in1k}
     \end{subfigure}
     \hfill
      \begin{subfigure}[b]{0.32\textwidth}
         \centering
         \includegraphics[width=\textwidth]{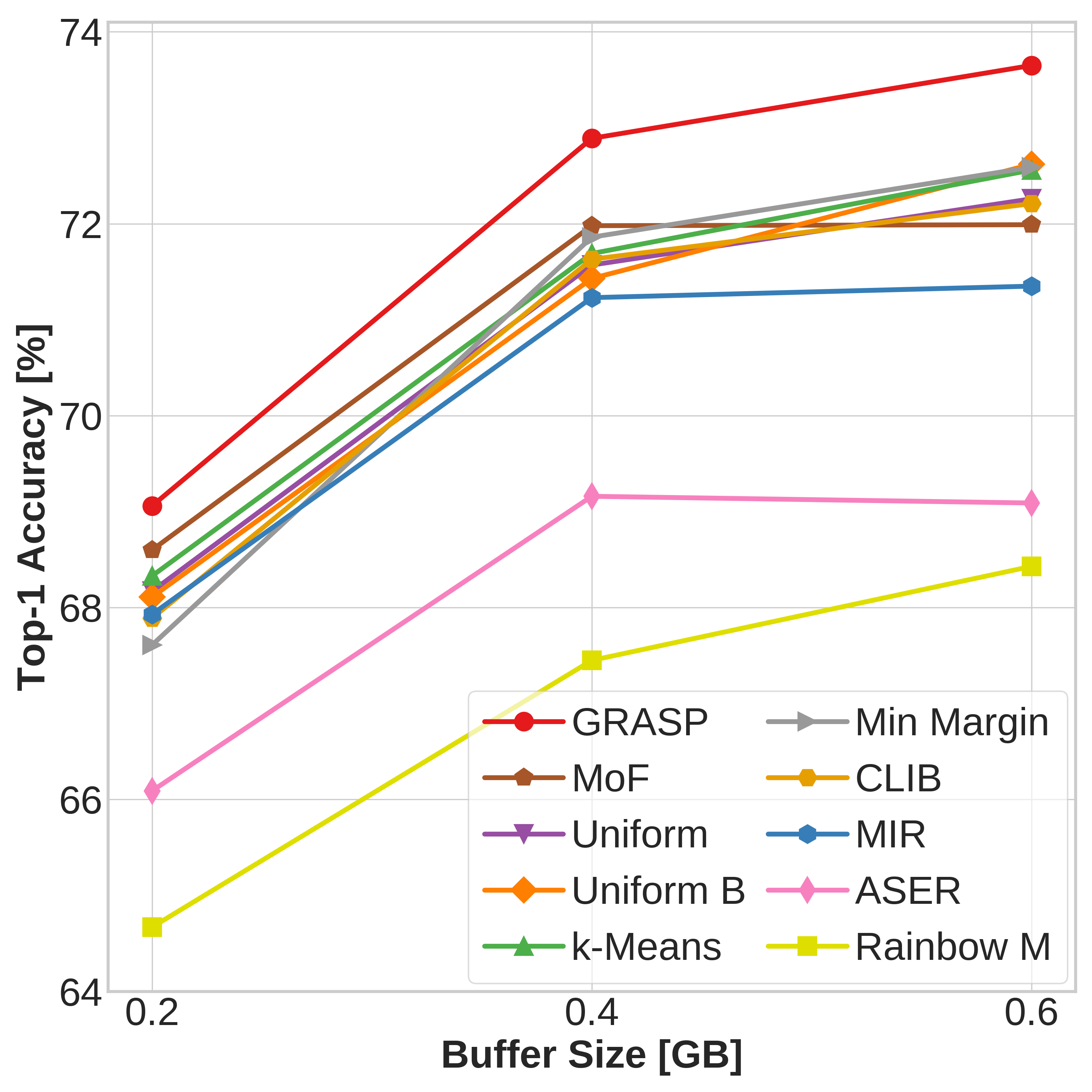}
         \caption{Varied Memory Constraints}
         \label{fig:memory}
     \end{subfigure}
     \hfill
      \begin{subfigure}[b]{0.32\textwidth}
         \centering
         \includegraphics[width=\textwidth]{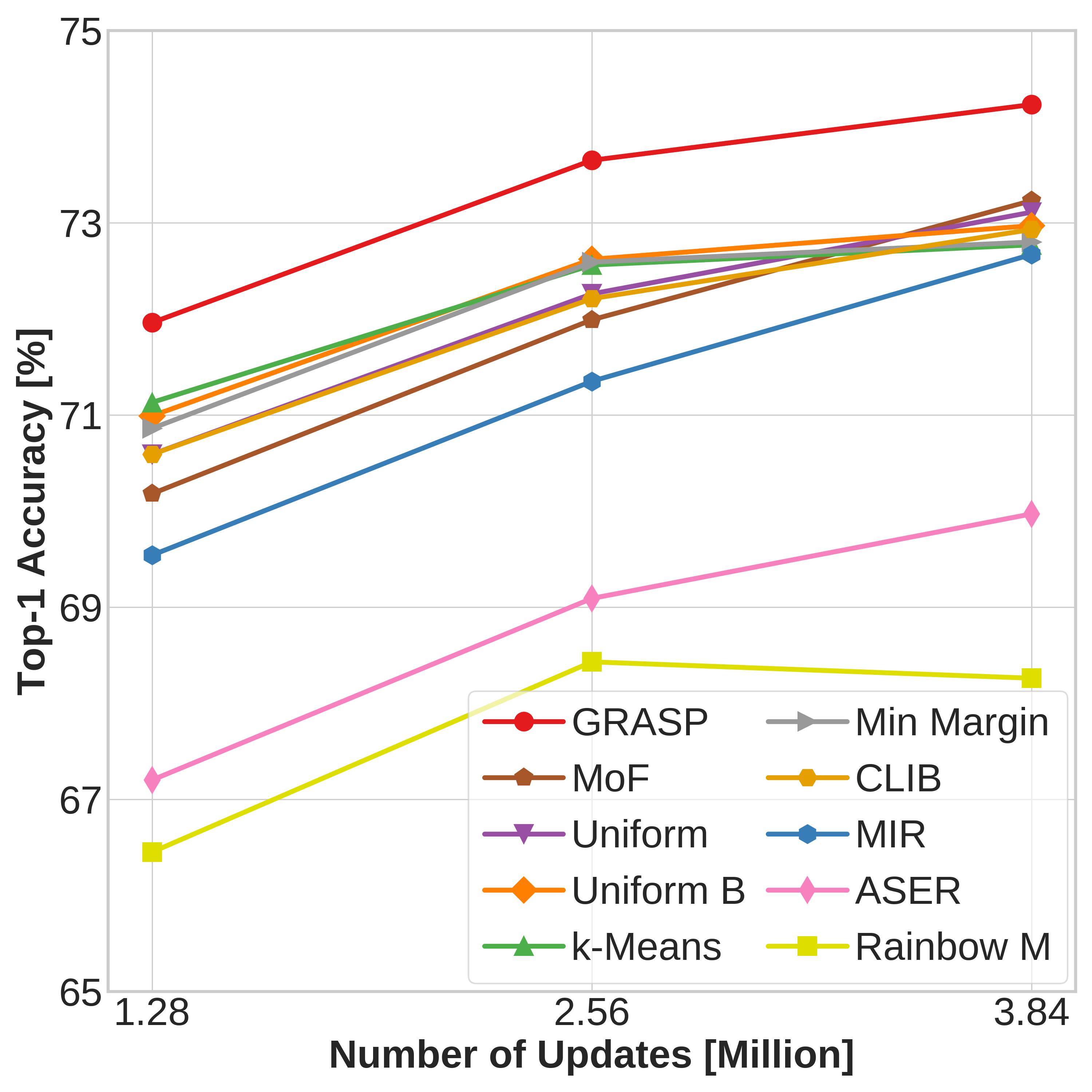}
         \caption{Varied Compute Constraints}
         \label{fig:compute}
     \end{subfigure}
     \hfill
   \caption{ (a) The ImageNet-1K learning curves of GRASP and uniform balanced policies in CIL using SIESTA. (b and c) The final accuracy of various rehearsal policies in CIL on ImageNet-300 using SIESTA and latent rehearsal.
   \label{fig:three_plots}}
   % \vspace{-1em}
\end{figure}

%\jc{prof suggests changing the subsection title to focus on "evaluating 14 rehearsal method"} \yh{done}
First, we analyze the performance of the 14 rehearsal methods including data-driven and model-based methods on ImageNet-300 using SIESTA with latent rehearsal during CIL. 
%\jc{"including 5 SoTA" could be removed since these sota do not outperform conventional ones; or say in their work their methods outperform others.} \jg{Done}
After pre-training MobileNet on the first 100 classes, the remaining 200 classes are learned in 4 rehearsal sessions (50 classes per rehearsal session).
%(50 classes and 1251 iterations per rehearsal session). 
%Results are summarized in 
As shown in Table~\ref{tab:replay_baseline_updated}, in both unbounded and bounded memory settings, GRASP achieves the highest final and average accuracy on all classes. In all criteria, GRASP outperforms uniform balanced rehearsal, which earlier works found was the most effective policy for large-scale datasets~\citep{prabhu2023computationally}. 
%\textcolor{magenta}{GRASP also outperforms other SoTA methods while providing significant computational benefits. In particular, GRASP is $2.4\times$, $5.3\times$, $11.8\times$, and $17.9\times$ faster than CLIB, MIR, Rainbow Memory, and ASER respectively in terms of training time.} 
Training time comparison is given in Fig.~\ref{fig:relative_train_time}.
%\paragraph{Impact of Memory and Compute Constraints.}
%\jc{Could be moved above; merge with 5.1} \yh{Done}
To validate that GRASP performs effectively under varied memory and compute constraints, we compare GRASP with competitive rehearsal policies i.e., MoF, uniform, uniform balanced, min margin, $k$-Means, CLIB, MIR, Rainbow Memory, and ASER under varied memory and compute constraints. As shown in Fig.~\ref{fig:memory} and~\ref{fig:compute}, %These experiments use settings from latent rehearsal experiments on ImageNet-300 in Sec.~\ref{sec:imagenet_subset_main}. 
GRASP consistently surpasses compared methods in all cases. In Appendix~\ref{sec.vis}, we also qualitatively compare GRASP with these competitive methods.

\begin{table}[t!]
  %\footnotesize
  \caption{ 
   \textbf{GRASP vs. Various Rehearsal Methods.} This uses latent rehearsal for CIL with SIESTA on \textbf{ImageNet-300}. $\mu_{A}$ denotes accuracy (\%) averaged over rehearsals, and $\alpha$ is the final accuracy (\%). Training time $T$ is in hours. %\textcolor{magenta}{We report mean and standard deviation of 3 runs exclusively for the competitive methods.}
   }
  \centering
     \begin{tabular}{c|cc|cc|c}
     \hline
     \multicolumn{1}{c|}{\textbf{Method}} &
     \multicolumn{2}{c|}{\textbf{Unbounded Memory}} &
     \multicolumn{2}{c}{\textbf{Bounded Memory}} & 
     \multicolumn{1}{|c}{\textbf{Time}} \\
      & $\mu_{A} \uparrow$ & $\alpha \uparrow$ & $\mu_{A} \uparrow$ & $\alpha \uparrow$ & $T \downarrow$ \\
      \hline
     Uniform & $77.46$ & $72.26$ & $75.36$ & $68.17$ & $0.30$ \\
     Min Confidence~\citep{gal2017deep} & $77.69$ & $72.42$ & $75.11$ & $67.79$ & $0.41$ \\
     Min Margin~\citep{scheffer2001active} & $77.51$ & $72.59$ & $75.20$ & $67.61$ & $0.41$ \\
     Max Loss~\citep{kawaguchi2020ordered} & $75.40$ & $68.89$ & $72.96$ & $64.55$ & $0.41$ \\
     Min Logit Dist~\citep{chaudhry2018riemannian} & $77.36$ & $72.07$ & $75.19$ & $67.85$ & $0.41$ \\
     $k$-Means~\citep{chaudhryER_2019} & $77.63$ & $72.56$ & $75.50$ & $68.33$ & $0.56$ \\
     Min Rehearsal~\citep{hayes2021selective} & $75.76$ & $69.76$ & $74.87$ & $67.37$ & $0.41$ \\
     MoF / Easy Biased~\citep{chaudhryER_2019} & $77.43$ & $71.99$ & $75.54$ & $68.60$ & $0.33$ \\
     Hard Biased & $76.90$ & $71.89$ & $74.75$ & $67.59$ & $0.33$ \\
     Uniform Balanced & $77.52$ & $72.62$ & $75.39$ & $68.11$ & $0.32$ \\
     %Uniform & $77.55$ \tiny $\pm 0.06$ & $72.45$ \tiny $\pm 0.22$ & $75.61$ \tiny $\pm 0.19$ & $68.39$ \tiny $\pm 0.26$ & $0.30$ \\
     %Min Margin~\citep{scheffer2001active} & $77.56$ \tiny $\pm 0.05$ & $72.42$ \tiny $\pm 0.25$ & $75.40$ \tiny $\pm 0.30$ & $68.09$ \tiny $\pm 0.35$ & $0.41$ \\
     %$k$-Means~\citep{chaudhryER_2019} & $77.72$ \tiny $\pm 0.10$ & $72.81$ \tiny $\pm 0.21$ & $75.74$ \tiny $\pm 0.17$ & $68.49$ \tiny $\pm 0.31$ & $0.56$ \\
     %MoF / Easy Biased~\citep{chaudhryER_2019} & $77.38$ \tiny $\pm 0.08$ & $72.14$ \tiny $\pm 0.11$ & $75.73$ \tiny $\pm 0.16$ & $68.70$ \tiny $\pm 0.18$ & $0.33$ \\
     %Uniform Balanced~\citep{prabhu2020gdumb} & $77.62$ \tiny $\pm 0.09$ & $72.67$ \tiny $\pm 0.12$ & $75.55$ \tiny $\pm 0.11$ & $68.34$ \tiny $\pm 0.17$ & $0.32$ \\
     %MIR~\citep{aljundi2019online} & $77.33$ \tiny $\pm 0.07$ & $71.63$ \tiny $\pm 0.25$ & $75.56$ \tiny $\pm 0.22$ & $68.27$ \tiny $\pm 0.27$ & $1.74$ \\
     %CLIB~\citep{koh2021online} & $77.56$ \tiny $\pm 0.09$ & $72.38$ \tiny $\pm 0.25$ & $75.40$ \tiny $\pm 0.15$ & $68.32$ \tiny $\pm 0.31$ & $0.78$ \\
     %\textbf{GRASP} (Ours) & $\mathbf{78.31}$ \tiny $\pm 0.10$ & $\mathbf{73.75}$ \tiny $\pm 0.09$ & $\mathbf{75.93}$ \tiny $\pm 0.14$ & $\mathbf{68.75}$ \tiny $\pm 0.22$ & $0.33$ \\
     Rainbow Memory~\citep{bang2021rainbow} & $74.93$ & $68.43$ & $72.73$ & $64.67$ & $3.89$ \\
     MIR~\citep{aljundi2019online} & $77.33$ & $71.35$ & $75.30$ & $67.93$ & $1.74$ \\
     CLIB~\citep{koh2021online} & $77.44$ & $72.21$ & $75.22$ & $67.89$ & $0.78$ \\
     ASER~\citep{shim2021online} & $75.16$ & $69.09$ & $73.79$ & $66.09$ & $5.91$ \\
     \hline
     %\cline{2-7}
     \textbf{GRASP} (Ours) & $\mathbf{78.39}$ & $\mathbf{73.65}$ & $\mathbf{76.12}$ & $\mathbf{69.06}$ & $0.33$ \\
     \hline
    \end{tabular}
  \label{tab:replay_baseline_updated}
\end{table} % imagenet-300 only
\begin{table}[t!]
  %\footnotesize
  \caption{ 
   \textbf{GRASP vs. Gradient-Based Methods.} This uses latent rehearsal for CIL with SIESTA on \textbf{ImageNet-150}. $\mu_{A}$ denotes accuracy (\%) averaged over rehearsals, and $\alpha$ is the final accuracy (\%). Training time $T$ is in hours. %Due to large computational cost, ImageNet-150 is used for gradient-based methods.
   }
  \centering
     \begin{tabular}{c|cc|cc|c}
     \hline
     \multicolumn{1}{c|}{\textbf{Method}} &
     \multicolumn{2}{c|}{\textbf{Unbounded Memory}} &
     \multicolumn{2}{c}{\textbf{Bounded Memory}} & 
     \multicolumn{1}{|c}{\textbf{Time}} \\
      & $\mu_{A} \uparrow$ & $\alpha \uparrow$ & $\mu_{A} \uparrow$ & $\alpha \uparrow$ & $T \downarrow$ \\
     \hline
     OCS~\citep{yoononline} & $74.93$ & $70.36$ & $74.13$ & $69.01$ & $7.34$ \\
     GSS~\citep{aljundi2019gradient} & $75.04$ & $70.25$ & $75.38$ & $70.47$ & $11.60$ \\
     Grad Matching~\citep{campbell2019automated} & $76.48$ & $72.20$ & $76.71$ & $72.36$ & $12.71$ \\
     \hline
     \textbf{GRASP} (Ours) & $\mathbf{77.75}$ & $\mathbf{73.96}$ & $\mathbf{77.88}$ & $\mathbf{73.71}$ & $0.13$ \\
     \hline
    \end{tabular}
  \label{tab:grad_baselines}
\end{table} % imagenet-150 only

\subsection{GRASP vs. Gradient-Based Methods}
\label{sec:grad_main}

We also compare GRASP with SoTA gradient-based methods e.g., OCS, GSS, and Grad Matching. 
It is computationally prohibitive to scale these methods (see Fig.~\ref{fig:relative_train_time}), for instance, Grad Matching requires $97\times$ more training time than GRASP to learn 50 ImageNet classes. 
Therefore we had to keep this comparison small scale with ImageNet-150 subset. After pre-training 100 classes, the next 50 classes are learned in 5 rehearsal sessions or tasks (10 classes per rehearsal). As shown in Table~\ref{tab:grad_baselines}, GRASP outperforms compared methods with significantly less training time.
%while achieving significantly higher computational efficiency. In particular, GRASP is $56.5\times$, $89.2\times$, and $97.8\times$ faster than OCS, GSS, and Grad Matching respectively.

%\input{tables/grad_baselines} % imagenet-150 only

\begin{table}[t]
  %\footnotesize
  \caption{ 
  \textbf{Latent Rehearsal Results for CIL with SIESTA on ImageNet-1K (3 runs).}
  %Here $\mu_{A}$ denotes accuracy (\%) on all seen classes averaged over rehearsal sessions and $\alpha$ is the final ImageNet-1K accuracy (\%).
  $\mu_{N}$, $\mu_{O}$, and $\mu_{A}$ denote accuracy on new, old, and all classes respectively averaged over rehearsals, and $\alpha$ is final ImageNet-1K accuracy.
   }
  \centering
  \resizebox{\linewidth}{!}{
     \begin{tabular}{l|cccc|cccc}
     \hline
     \multicolumn{1}{c|}{\textbf{Method}} &
     \multicolumn{4}{c|}{\textbf{Unbounded Memory}} &
     \multicolumn{4}{c}{\textbf{Bounded Memory}} \\
     %& $\mu_{A} \uparrow$ & $\alpha \uparrow$ & $\mu_{A} \uparrow$ & $\alpha \uparrow$ \\
     & $\mu_{N} \uparrow$ & $\mu_{O} \uparrow$ & $\mu_{A} \uparrow$ & $\alpha \uparrow$ & $\mu_{N} \uparrow$ & $\mu_{O} \uparrow$ & $\mu_{A} \uparrow$ & $\alpha \uparrow$ \\
     %\hline
     %Offline Model & --- & --- & --- & $62.12$ & --- & --- & --- & $62.12$ \\
     \hline
     %Uniform Bal & $66.82$ & $69.33$ & $68.88$ & $60.68$ & $67.20$ & $69.26$ & $68.88$ & $60.52$ \\ % seed 1444
     %Uniform Bal & $67.15$ & $69.36$ & $68.96$ & $60.80$ & $66.99$ & $69.22$ & $68.81$ & $60.25$ \\ % seed 1221
     %Uniform Bal & $66.96$ & $69.36$ & $68.93$ & $60.79$ & $67.02$ & $69.34$ & $68.95$ & $60.27$ \\ % seed 1993
     %%%%%%%
     %GRASP & $68.10$ & $70.41$ & $70.03$ & $62.14$ & $68.26$ & $70.40$ & $70.01$ & $61.66$ \\ % seed 1444
     %GRASP & $67.98$ & $70.33$ & $69.93$ & $62.07$ & $68.15$ & $70.40$ & $70.01$ & $61.79$ \\ % seed 1221
     %GRASP & 68.30$ & $70.46$ & $70.11$ & $61.93$ & $68.25$ & $70.30$ & $69.96$ & $61.99$ \\ % seed 1993
     %%%%%
     Uniform B & $66.98$ \tiny $\pm 0.14$ & $69.35$ \tiny $\pm 0.01$ & $68.92$ \tiny $\pm 0.03$ & $60.76$ \tiny $\pm 0.05$ & $67.07$ \tiny $\pm 0.09$ & $69.27$ \tiny $\pm 0.05$ & $68.88$ \tiny $\pm 0.06$ & $60.35$ \tiny $\pm 0.12$ \\ %avg+-std
     \textbf{GRASP} & $\mathbf{68.13}$ \tiny $\pm 0.13$ & $\mathbf{70.40}$ \tiny $\pm 0.05$ & $\mathbf{70.02}$ \tiny $\pm 0.07$ & $\mathbf{62.05}$ \tiny $\pm 0.09$ & $\mathbf{68.22}$ \tiny $\pm 0.05$ & $\mathbf{70.37}$ \tiny $\pm 0.05$ & $\mathbf{70.00}$ \tiny $\pm 0.02$ & $\mathbf{61.81}$ \tiny $\pm 0.14$ \\ % avg+-std
     %%%%%
     %Uniform Bal & $68.92$ \tiny $\pm 0.03$ & $60.76$ \tiny $\pm 0.05$ & $68.88$ \tiny $\pm 0.06$ & $60.35$ \tiny $\pm 0.12$ \\ %avg+-std
     %\textbf{GRASP} & $\mathbf{70.02}$ \tiny $\pm 0.07$ & $\mathbf{62.05}$ \tiny $\pm 0.09$ & $\mathbf{70.00}$ \tiny $\pm 0.02$ & $\mathbf{61.81}$ \tiny $\pm 0.14$ \\ % avg+-std
     \hline
    \end{tabular}}
  \label{tab:replay_comp}
\end{table}
 % ImageNet-1K Results (Latent Rehearsal)

\begin{table}[t!]
  %\footnotesize
  \caption{
  \textbf{Veridical Rehearsal Results for CIL with SIESTA on ImageNet-1K (3 runs).}
  %Here $\mu_{A}$ denotes accuracy (\%) on all seen classes averaged over rehearsal sessions and $\alpha$ is the final ImageNet-1K accuracy (\%).
  $\mu_{N}$, $\mu_{O}$, and $\mu_{A}$ denote accuracy on new, old, and all classes respectively averaged over rehearsals, and $\alpha$ is final ImageNet-1K accuracy.
  }
  \centering
     \resizebox{\linewidth}{!}{
     \begin{tabular}{l|cccc|cccc}
     \hline
     \multicolumn{1}{c|}{\textbf{Method}} &
     \multicolumn{4}{c|}{\textbf{Unbounded Memory}} &
     \multicolumn{4}{c}{\textbf{Bounded Memory}} \\
     % & $\mu_{A} \uparrow$ & $\alpha \uparrow$ & $\mu_{A} \uparrow$ & $\alpha \uparrow$ \\
     & $\mu_{N} \uparrow$ & $\mu_{O} \uparrow$ & $\mu_{A} \uparrow$ & $\alpha \uparrow$ & $\mu_{N} \uparrow$ & $\mu_{O} \uparrow$ & $\mu_{A} \uparrow$ & $\alpha \uparrow$ \\
     %\hline
     %Joint & --- & --- & --- & $62.12$ & --- & --- & --- & $62.12$ \\
     %%%%%%%%%%%%%% Uniform Balalnced %%%%%%%%%%%
     %Uniform B & 68.27 & 67.46 & 67.42 & 54.03 & 63.48 & 61.67 & 61.82 & 46.05 \\ % seed 1444
     %Uniform B & 68.33 & 67.59 & 67.51 & 55.08 & 63.64 & 61.60 & 61.82 & 46.34 \\ % seed 1221
     %Uniform B & 69.20 & 70.81 & 70.50 & 59.49 & 63.62 & 62.86 & 62.86 & 47.98 \\ % seed 1993
     %%%%%%%%%%%%%% GRASP %%%%%%%%%%%
     %GRASP & 69.62 & 72.54 & 72.10 & 63.57 & 63.98 & 63.32 & 63.32 & 49.11 \\ % seed 1444
     %GRASP & 69.60 & 72.69 & 72.22 & 63.58 & 64.17 & 63.36 & 63.41 & 49.38 \\ % seed 1221
     %GRASP & 70.33 & 72.84 & 72.40 & 64.45 & 64.06 & 63.40 & 63.41 & 49.31 \\ % seed 1993
     \hline
     Uniform B & $68.60$ \tiny $\pm 0.42$ & $68.62$ \tiny $\pm 1.55$ & $68.48$ \tiny $\pm 1.43$ & $56.20$ \tiny $\pm 2.37$ & $63.58$ \tiny $\pm 0.07$ & $62.04$ \tiny $\pm 0.58$  & $62.17$ \tiny $\pm 0.49$ & $46.79$ \tiny $\pm 0.85$ \\
     \textbf{GRASP} & $\mathbf{69.85}$ \tiny $\pm 0.34$ & $\mathbf{72.69}$ \tiny $\pm 0.12$ & $\mathbf{72.24}$ \tiny $\pm 0.12$ & $\mathbf{63.87}$ \tiny $\pm 0.41$ & $\mathbf{64.07}$ \tiny $\pm 0.08$ & $\mathbf{63.36}$ \tiny $\pm 0.03$ & $\mathbf{63.38}$ \tiny $\pm 0.04$ & $\mathbf{49.27}$ \tiny $\pm 0.11$ \\
     %\hline
     %Uniform Bal & $68.48$ \tiny $\pm 1.43$ & $56.20$ \tiny $\pm 2.37$ & $62.17$ \tiny $\pm 0.49$ & $46.79$ \tiny $\pm 0.85$ \\
     %\textbf{GRASP} & $\mathbf{72.24}$ \tiny $\pm 0.12$ & $\mathbf{63.87}$ \tiny $\pm 0.41$ & $\mathbf{63.38}$ \tiny $\pm 0.04$ & $\mathbf{49.27}$ \tiny $\pm 0.11$ \\
     \hline
    \end{tabular}}
  \label{tab:veridical}
\end{table}
 % ImageNet-1K Results (Veridical Rehearsal)
\begin{table}[t]
  %\footnotesize
  \caption{
  \textbf{DERpp \& GDumb (3 runs).}
   %Comparison among GRASP and uniform balanced 
   Comparison among rehearsal policies when combined with DERpp and GDumb in offline CIL on \textbf{ImageNet-1K}. %%%Both compute and memory constraints are imposed.
   %Uniform balanced policy is referred as Uniform$\dag$. 
   %Here $\mu_{N}$, $\mu_{O}$, and $\mu_{A}$ denote accuracy ($\%$) on new, old, and all classes respectively averaged over rehearsal sessions. And $\alpha$ is the final accuracy ($\%$) on all classes.
   $\dag$ and $\ddag$ denote variants that use uniform balanced and GRASP respectively.
   }
  \centering
  \resizebox{\linewidth}{!}{
     \begin{tabular}{l|cccc|cccc}
     \hline
     \multicolumn{1}{c|}{\textbf{Method}} &
     \multicolumn{4}{c|}{\textbf{Latent Rehearsal}} &
     \multicolumn{4}{c}{\textbf{Veridical Rehearsal}} \\
     
      & $\mu_{N} \uparrow$ & $\mu_{O} \uparrow$ & $\mu_{A} \uparrow$ & $\alpha \uparrow$ & $\mu_{N} \uparrow$ & $\mu_{O} \uparrow$ & $\mu_{A} \uparrow$ & $\alpha \uparrow$ \\
     %\hline
     %Offline Model & --- & --- & --- & $62.12$ & --- & --- & --- & $62.12$ \\
     %%%%%%% DERpp %%%%%%%%
     %uniform & 76.28 & 64.25 & 65.95 & 53.90 & 58.71 & 61.51 & 60.97 & 44.15 \\ % seed 1444
     %uniform & 76.41 & 64.60 & 66.25 & 54.04 & 50.40 & 61.72 & 60.09 & 43.32 \\ % seed 1221
     %uniform & 76.24 & 64.36 & 66.03 & 53.74 & 62.31 & 63.12 & 62.76 & 47.77 \\ % seed 1993
     %grasp & 77.48 & 65.07 & 66.83 & 54.49 & 52.10 & 63.19 & 61.34 & 47.93 \\ % seed 1444
     %grasp & 77.49 & 65.01 & 66.79 & 54.29 & 52.47 & 63.15 & 61.35 & 48.00 \\ % seed 1221
     %grasp & 77.66 & 65.04 & 66.84 & 54.64 & 59.67 & 63.91 & 63.04 & 49.78 \\ % seed 1993
     \hline
     DERpp$\dag$ & $76.31$ \tiny $\pm 0.07$ & $64.40$ \tiny $\pm 0.15$ & $66.08$ \tiny $\pm 0.13$ & $53.89$ \tiny $\pm 0.12$ & $\mathbf{57.14}$ \tiny $\pm 4.99$ & $62.12$ \tiny $\pm 0.71$ & $61.27$ \tiny $\pm 1.11$ & $45.08$ \tiny $\pm 1.93$ \\
     \textbf{DERpp}$\ddag$ & $\mathbf{77.54}$ \tiny $\pm 0.08$ & $\mathbf{65.04}$ \tiny $\pm 0.02$ & $\mathbf{66.82}$ \tiny $\pm 0.02$ & $\mathbf{54.47}$ \tiny $\pm 0.14$ & $54.75$ \tiny $\pm 3.48$ & $\mathbf{63.42}$ \tiny $\pm 0.35$ & $\mathbf{61.91}$ \tiny $\pm 0.80$ & $\mathbf{48.57}$ \tiny $\pm 0.86$ \\
     %%%%%%% GDumb %%%%%%%%
     %uniform & 67.71 & 69.71 & 69.34 & 61.09 & 62.16 & 62.61 & 62.45 & 46.87 \\ % seed 1444
     %uniform & 67.75 & 69.92 & 69.51 & 60.95 & 61.94 & 62.53 & 62.36 & 46.70 \\ % seed 1221
     %uniform & 67.70 & 69.76 & 69.39 & 61.11 & 63.10 & 63.39 & 63.25 & 48.77 \\ % seed 1993
     %grasp & 69.09 & 71.00 & 70.64 & 62.78 & 63.53 & 64.14 & 63.94 & 50.01 \\ % seed 1444
     %grasp & 69.00 & 70.74 & 70.43 & 62.78 & 63.74 & 64.16 & 64.02 & 50.16 \\ % seed 1221
     %grasp & 69.03 & 70.78 & 70.47 & 62.63 & 63.76 & 64.12 & 64.00 & 49.93 % seed 1993
     \hline
     GDumb$\dag$ & $67.72$ \tiny $\pm 0.02$ & $69.80$ \tiny $\pm 0.09$ & $69.41$ \tiny $\pm 0.07$ & $61.05$ \tiny $\pm 0.07$ & $62.40$ \tiny $\pm 0.50$ & $62.84$ \tiny $\pm 0.39$ & $62.69$ \tiny $\pm 0.40$ & $47.45$ \tiny $\pm 0.94$ \\
     \textbf{GDumb}$\ddag$ & $\mathbf{69.04}$ \tiny $\pm 0.04$ & $\mathbf{70.84}$ \tiny $\pm 0.11$ & $\mathbf{70.51}$ \tiny $\pm 0.09$ & $\mathbf{62.73}$ \tiny $\pm 0.07$ & $\mathbf{63.68}$ \tiny $\pm 0.10$ & $\mathbf{64.14}$ \tiny $\pm 0.02$ & $\mathbf{64.00}$ \tiny $\pm 0.03$ & $\mathbf{50.03}$ \tiny $\pm 0.10$ \\
     \hline
    \end{tabular}} 
  \label{tab:other_cl}
\end{table} % Other CL methods: DER++ and GDumb

%%%%%%%%%%%%%%%%%%%%%%%%%%%%%%%%%%%%%
\subsection{ImageNet-1K Experiments}
\label{sec:imagenet1k_main}

Having shown that under the same computational budget, GRASP achieves SoTA accuracy with little computational overhead compared to uniform when combined with SIESTA, we next turn to assessing GRASP's abilities on ImageNet-1K under a variety of scenarios: latent rehearsal, veridical rehearsal, IID CL, and generalization to other algorithms beyond SIESTA. As a baseline, we use balanced uniform in these experiments. %\ck{polish this, but we need to introduce what we are doing - we should not say the bit about computational efficiency as many of them are only slightly more expensive} \yh{done}
In Appendix~\ref{sec.vis}, we also analyze the performance improvements of GRASP over uniform balanced in various ImageNet-1K experiments.

%\paragraph{SIESTA with Latent Rehearsal.}
%\paragraph{Latent Rehearsal.}
\textbf{Latent Rehearsal.}
ImageNet-1K results for CIL with SIESTA using latent rehearsal are given in Table~\ref{tab:replay_comp}. GRASP consistently outperforms uniform balanced across criteria in both unbounded and bounded memory settings. 
Learning curves for GRASP and uniform balanced latent rehearsal are given in Fig.~\ref{fig:in1k}. GRASP (latent) achieves higher accuracy than uniform balanced (latent) in all rehearsal sessions (100 ImageNet classes per rehearsal session).
GRASP provides $40\%$ and $36\%$ speedups in terms of compute and training time respectively (see Fig.~\ref{fig:compute_time}).
%\paragraph{Impact of Memory and Compute Constraints.}
Additionally, in Table~\ref{tab:compute_memory_analysis} we compare GRASP with uniform balanced on ImageNet-1K under varied compute and memory constraints. %using settings from latent rehearsal experiments on ImageNet-1K in Sec.~\ref{sec:imagenet1k_main}. 
Under all circumstances, GRASP consistently exceeds uniform balanced.
Using McNemar's test, we compare the predictive accuracy of GRASP and uniform balanced and find that they are significantly different ($P<0.001$) in all cases.

%\paragraph{Veridical Rehearsal.} 
\textbf{Veridical Rehearsal.} 
To assess if GRASP is effective for veridical rehearsal instead, we compared GRASP to uniform balanced rehearsal with a variant of SIESTA that stores raw images. CIL results on ImageNet-1K are given in Table~\ref{tab:veridical}, GRASP persistently outperforms uniform balanced baseline in both unbounded and bounded memory settings. 
In terms of final ImageNet-1K accuracy, GRASP exceeds uniform balanced by absolute $7.67\%$ (unbounded memory) and $2.48\%$ (bounded memory).
Fig.~\ref{fig:in1k} shows learning curves. GRASP (veridical) obtains higher accuracy than uniform balanced (veridical) in all rehearsal sessions (100 ImageNet classes per rehearsal 
session). Additionally, we show ImageNet-1K curves for old and new tasks in Fig.~\ref{fig:old_new}.
We compare the final predictions of GRASP and uniform balanced using McNemar's test and find that they are significantly different ($P<0.001$) in all experiments.

%%%%%%%%%%%%%%%%%%%%%%%%%%%%%%%%%%%%%%%%%%%%%%%%%%
%\paragraph{Continual IID Learning.}
\textbf{Continual IID Learning.}
An ideal rehearsal policy should excel regardless of distribution. CIL is an extreme adversarial setting where catastrophic forgetting is severe. Here we consider the other extreme, IID CL, where catastrophic forgetting is minimal~\citep{hayes2018new}. 
In IID CL settings, we conduct bounded-memory experiments (3 runs) on ImageNet-1K using SIESTA with latent rehearsal where each task contains 128K samples from randomly sampled classes. Other details adhere to the CIL's ImageNet-1K bounded memory setting.
%An ideal rehearsal policy should excel regardless of distribution.
%\jc{move motivation to the beginning.} \yh{Done}
%For IID CL experiments using SIESTA with latent rehearsal, 
GRASP achieves higher accuracy ($\mu_{A}=61.49 \pm0.02$ and $\alpha=63.22 \pm0.09$) than uniform balanced ($\mu_{A}=60.32 \pm0.06$ and $\alpha=61.52 \pm0.03$). %, where $\mu$ and $\alpha$ are average and final accuracy respectively. 
McNemar's test shows a significant difference ($P<0.001$) between GRASP and uniform balanced policies.
%, where $\mu$ and $\alpha$ denote average accuracy ($\%$) over rehearsals and final accuracy ($\%$), respectively.

%\paragraph{Experiments with GDumb and DERpp.}
\textbf{Experiments with GDumb and DERpp.}
%\jc{Motivation: To ensure GRASP not only works with SIESTA...} \yh{Done}
To validate that GRASP shows effectiveness for other rehearsal-based CL methods beyond SIESTA, we combined GRASP with two commonly used offline CL methods that use rehearsal: GDumb and DERpp for both the latent and veridical rehearsal settings.
%For both the latent and veridical rehearsal settings, we combined GRASP with two commonly used offline CL methods that use rehearsal: GDumb and DERpp. 
These experiments were done with the same MobileNetV3-L architecture used by SIESTA, which was pre-trained on the first 100 ImageNet classes. 
Compute and memory constraints are imposed. On CIL experiments with ImageNet-1K, %we found that 
GRASP outperformed the uniform balanced, as shown in Table~\ref{tab:other_cl}. 
McNemar's test comparing final predictions of GRASP and uniform balanced reveals a significant difference ($P<0.001$) between them in all conditions.

\subsection{Additional Experiments}
\label{sec:additional_exp}
% The remaining experiments compare GRASP to uniform sampling. 

\begin{table}[t!]
  %\footnotesize
  \caption{ 
  \textbf{Compute and Memory Constraints Analysis for CIL with SIESTA on ImageNet-1K.}
  %GRASP vs. uniform balanced under various compute and memory constraints in CIL on \textbf{ImageNet-1K} with SIESTA using latent rehearsal. 
  Buffer size and number of updates are reported in GB and million respectively. Reported is the final accuracy ($\%$) on $1000$ classes.}
  \centering
     \begin{tabular}{c|ccc|ccc}
     \hline
     \multicolumn{1}{c|}{\textbf{Method}} &
     \multicolumn{3}{c|}{\textbf{Updates (M)}} &
     \multicolumn{3}{c}{\textbf{Buffer (GB)}} \\
     
      & $0.76$ & $1.02$ & $1.53$ & $0.75$ & $1.51$ & $2.01$ \\
     \hline
     Uniform Bal & $59.10$ & $60.28$ & $61.10$ & $57.13$ & $60.27$ & $60.79$ \\
     \textbf{GRASP} & $\mathbf{60.00}$ & $\mathbf{61.36}$ & $\mathbf{62.66}$ & $\mathbf{58.44}$ & $\mathbf{62.00}$ & $\mathbf{61.93}$ \\
     \hline
    \end{tabular}
  \label{tab:compute_memory_analysis}
\end{table}
 % compute and memory comparison among GRASP and uniform

\begin{wrapfigure}[14]{t}{0.5\textwidth}
    % \begin{figure}[t]
    \vspace{-1.2cm}
       \includegraphics[width=\linewidth]{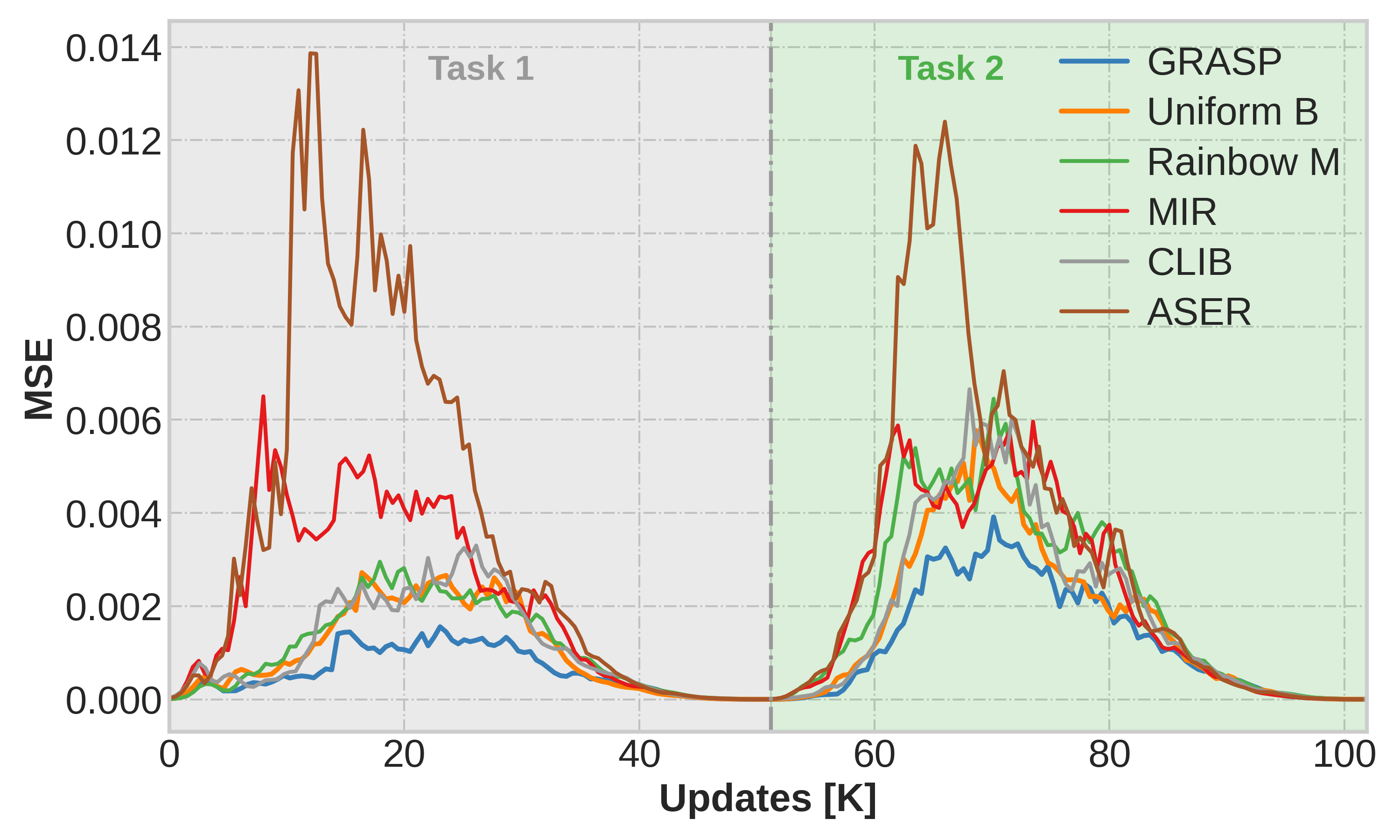}
       \vspace{-0.75cm}
       \caption{The representation drift of old classes while learning new classes in a stream of two tasks denoted by background colors. GRASP reduces representation drift.
       }
       \label{fig:drift}
    % \end{figure}
\end{wrapfigure}

%\paragraph{GRASP Reduces Representation Drift.}
%\textbf{GRASP Reduces Representation Drift.}
\textbf{Why is GRASP More Effective?}
%\jc{Needs motivation. This experiment explains why GRASP works. Representation drift is an issue for efficient CL. To test this we ... } \yh{Done}
An efficient rehearsal policy should not perturb previously learned representations otherwise training overhead increases with an increase in representation drift.
While learning new classes, the representations of old classes abruptly change and drift over time~\citep{caccia2021new}. This abrupt change in old representations causes catastrophic forgetting of old knowledge and is difficult to correct without longer training. 
As shown in Fig.~\ref{fig:drift}, existing methods exhibit higher representation drift. These methods mostly prioritize difficult samples, for instance, MIR and ASER select samples with maximum interference. Consequently, the old representations are excessively perturbed especially in the early stage of rehearsal indicated by the sharp rise in MSE.
On the contrary, GRASP reduces representation drift by learning from subsets of increasing difficulty levels. 
A quantitative comparison is also given in Table~\ref{tab:comp_dirft}. We see that compared to other competitive methods, GRASP achieves the lowest drift in all criteria.
Following~\citet{caccia2021new}, we measure the representation drift of an old sample $X$ at each training iteration $t$ as $\left\| f_{\theta_{t}}(X) - f_{\theta_{t+1}}(X) \right\|$ where $f_{\theta}$ denotes model parameters.
See Appendix~\ref{sec:drift_detail} for additional implementation details. 
%A quantitative comparison is given in Appendix~\ref{sec.drift_results}.

\begin{table}[t!]
  %\footnotesize
  \caption{
  \textbf{Representation drift in two tasks setup for CIL with SIESTA.}
  %GRASP vs. uniform balanced under various compute and memory constraints in CIL on \textbf{ImageNet-1K} with SIESTA using latent rehearsal. 
  We compare GRASP with other state-of-the-art policies in terms of representation drift on old tasks. %Each task consists of 10 ImageNet classes.
  Here, $\beta$ and $\phi$ denote AUC drift and average drift over training iterations respectively.
  %Buffer size and number of updates are reported in GB and million respectively. Reported is the final accuracy ($\%$) on $1000$ classes.
  }
  \centering
     \begin{tabular}{c|cc|cc}
     \hline
     \multicolumn{1}{c|}{\textbf{Method}} &
     \multicolumn{2}{c|}{\textbf{Task 1}} &
     \multicolumn{2}{c}{\textbf{Task 2}} \\
     & $\beta \downarrow$ & $\phi \downarrow$ & $\beta \downarrow$ & $\phi \downarrow$ \\
     \hline
     Rainbow Memory~\citep{bang2021rainbow} & $0.1100$ & $0.0010$ & $0.2148$ & $0.0021$ \\
     MIR~\citep{aljundi2019online} & $0.2192$ & $0.0022$ & $0.2273$ & $0.0022$ \\
     CLIB~\citep{koh2021online} & $0.1136$ & $0.0011$ & $0.1920$ & $0.0019$ \\
     ASER~\citep{shim2021online} & $0.3781$ & $0.0038$ & $0.3156$ & $0.0030$ \\
     Uniform Balanced & $0.1039$ & $0.0010$ & $0.1706$ & $0.0017$ \\
     \hline
     \textbf{GRASP} & $\mathbf{0.0600}$ & $\mathbf{0.0006}$ & $\mathbf{0.1275}$ & $\mathbf{0.0013}$ \\
     \hline
    \end{tabular}
  \label{tab:comp_dirft}
\end{table}

%\paragraph{Continual Text Classification.}
\textbf{Continual Text Classification.}
Using IDBR, we evaluate GRASP and uniform rehearsal policies in continual text classification. They perform task incremental learning with various task sequences based on 5 datasets (AG News, Yelp, DBPedia, Amazon, and Yahoo! Answer). Compute and memory constraints are imposed. Performance is averaged over 3 runs.
As shown in Table~\ref{tab:comp_nlp} in Appendix~\ref{sec.text_classification}, GRASP outperforms uniform in 5 out of 6 task sequences. Performance gains by GRASP align with the ones of IDBR~\citep{huang2021continual} in the same benchmark datasets.

%\paragraph{Long-Tailed Recognition.}
\textbf{Long-Tailed Recognition.}
%\jc{Motivation: some algorithm only works for a certain distribution, cite xxx.} \yh{Done}
Besides balanced data streams, a rehearsal policy should also work for long-tailed data streams since real-world data distributions are often imbalanced and long-tailed. 
%Here we analyze GRASP and uniform balanced rehearsal policies for CIL with Places-LT-365. 
In both unbounded and bounded memory settings, GRASP exceeds uniform balanced in CIL on Places-LT-365 (see Appendix~\ref{sec:lt_exp}).

\textbf{Vision Transformer Results.}
%\textbf{Vision Transformer Results.}
To examine GRASP's efficacy in a ViT architecture, we conduct CL experiments using MobileViT-small with SIESTA. 
%We compare GRASP with uniform balanced rehearsal in CIL on ImageNet-300. 
GRASP outperforms uniform balanced in CIL on ImageNet-300 (see Appendix~\ref{sec:vit}).

\section{Conclusion}
We showed that GRASP is a highly effective rehearsal policy compared to others on both large-scale image and NLP datasets. GRASP is effective for both latent and veridical rehearsal, and it works for multiple data distributions. GRASP is the first method to outperform balanced uniform for CIL on ImageNet-1K. We found that GRASP is more effective than other policies under a range of compute and memory constraints. %GRASP also showed effectiveness in reducing representation drift and preserving prior knowledge.

We focused on rehearsal policies, however, in future work, it would be interesting to examine the use of GRASP for buffer maintenance. We primarily focused on classification tasks to compare GRASP with the majority of existing sample selection policies that were originally designed for classification tasks.
Besides classification tasks, GRASP can be explored in other computer vision and NLP tasks, including continual object detection~\citep{acharya2020rodeo}. 
However, a suitable hardness score would have to be designed for other tasks since the distance to the class prototype is only appropriate for classification.
% GRASP can also be applied in relevant areas such as dataset pruning, active learning and unsupervised learning for efficiency purposes.
We studied GRASP with a fixed compute budget i.e., pre-defined fixed training steps. Future work could explore dynamically adapting the amount of training during rehearsal where the DNN stops early after achieving maximum performance.

While periodic retraining is currently the industry standard for updating DNNs, we believe GRASP is an important step toward supplanting this extremely computationally expensive process with much more efficient CL methods, and therefore reducing the carbon footprint from training models~\citep{wu2022sustainable}. Likewise, GRASP can be used to make on-device CL more efficient, where both compute and memory are heavily constrained~\citep{hayes2022online}.

\subsubsection*{Acknowledgments}
We thank Tyler Hayes for comments on an early version of the manuscript. This work was supported in part by NSF awards \#1909696, \#2326491, and \#2125362. The views and conclusions contained herein are those of the authors and should not be interpreted as representing the official policies or endorsements of any sponsor.

%\section{Rebuttal Modifications}
%When making changes to your paper during the rebuttal period, please use the \verb+EasyReview+ package to indicate what text has been \add{added}, \remove{removed} or \replace{replaced}{replaced}. Other potentially useful commands are referenced \href{http://mirrors.ibiblio.org/CTAN/macros/latex/contrib/easyreview/doc/easyReview.pdf}{here}.

\bibliography{collas2024_conference}
\bibliographystyle{collas2024_conference}

\clearpage

\appendix
%\section{Appendix}
%You may include other additional sections here.

\begin{center}
    {\Large{\textbf{Appendix}}}
\end{center}

\section{Overview of GRASP}

\begin{figure}[h]
  \centering
   \includegraphics[width=0.5\linewidth]{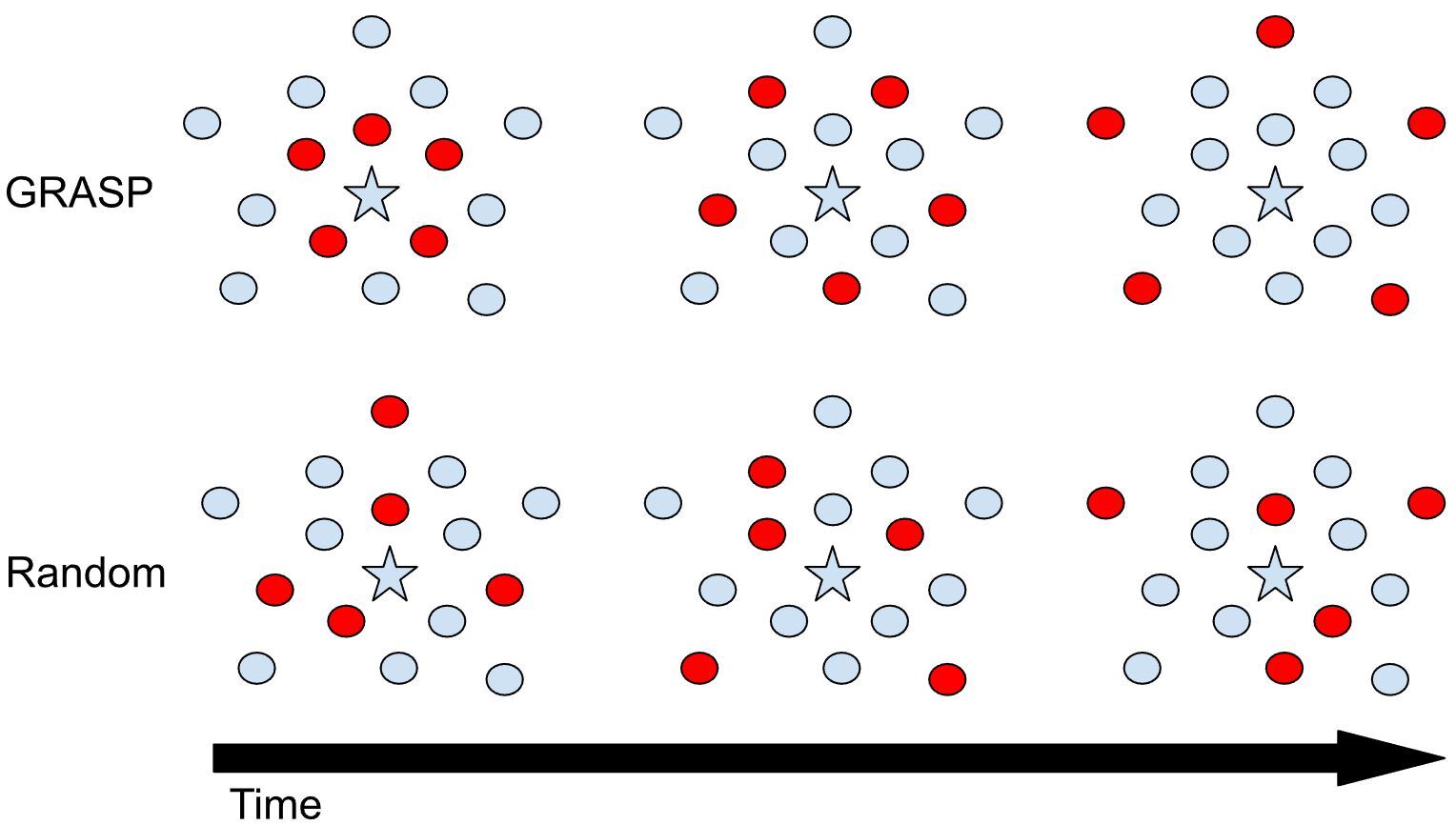}
   \caption{Overview of GRASP and Random Rehearsal Policies. Class mean is denoted by star. Selected samples are indicated by red circle.
   }
   \label{fig:grasp_vis}
\end{figure}

We illustrate how GRASP works compared to uniform random policy in Fig.~\ref{fig:grasp_vis}. We see that GRASP initially selects the most prototypical (representative) samples near the class mean and progressively selects less prototypical samples far from the class mean. Thus GRASP varies difficulty level to facilitate faster convergence compared to uniform random policy.

 % Fig showing how GRASP works

\section{Implementation Details}
\label{sec:implement}

%\textbf{MobileNet/ImageNet-1K Experiments.}
\subsection{MobileNet/ImageNet-1K Experiments}
For all MobileNet/ImageNet-1K experiments in Sec.~\ref{sec:imagenet1k_main}, we use SGD optimizer and OneCycle learning rate (LR) scheduler~\citep{smith2017super}. We set weight decay and momentum to $10^{-5}$ and $0.9$ respectively.
For latent rehearsal experiments, we use an initial LR of $1.6$ in the last layer with layer-wise LR reduction by a factor of $0.99$ for earlier layers and mini-batch size $512$.
For veridical rehearsal experiments, we set initial LR to $0.4$ for mini-batch size $256$ and use similar layer-wise LR reduction as before.

For MobileNet's base-initialization (pre-training) on 100 ImageNet classes, we adopt pre-trained weights from SIESTA. For additional details about pre-training, we refer readers to SIESTA paper~\citep{harun2023siesta}. We also use the same ImageNet-1K data ordering as used in SIESTA.
We configure OPQ to use 8 codebooks of size 256. OPQ is also trained on the same 100 ImageNet classes as used for pre-training MobileNet and kept fixed during the CL phase. OPQ is only used for latent rehearsal methods.

After base-initialization on 100 ImageNet classes, models continually learn the remaining 900 ImageNet classes in 9 rehearsal sessions (100 classes per rehearsal session). We set the number of iterations per rehearsal session to $2502$ for mini-batch size $512$ in latent rehearsal experiments.
Whereas, in veridical rehearsal experiments, we set the number of iterations per rehearsal session to $5004$ for a mini-batch size $256$.

All algorithms e.g., SIESTA, GDumb, and DERpp use the same settings e.g., the same hyperparameters and the same pre-trained MobileNet architecture.
For latent variants of GDumb and DERpp, we use the same configurations as SIESTA such as identical frozen earlier layers, identical plastic layers, the same pre-trained MobileNet architecture, and the same pre-trained OPQ model.

In all cases including both latent and veridical rehearsals, all methods use the same pre-trained MobileNet architecture and the same base initialization phase.
We do not apply image augmentation in any experiment to solely focus on rehearsal policy without the influence of other variables.

%\textbf{MobileNet/ImageNet-300 Experiments.}
\subsection{MobileNet/ImageNet-300 Experiments}
For MobileNet/ImageNet-300 experiments in Sec.~\ref{sec:imagenet_subset_main}, we use the same settings as aforementioned ImageNet-1K experiments e.g., hyperparameters, OPQ, optimizer, LR scheduler, and pre-trained MobileNet architecture.
After base-initialization on 100 ImageNet classes, models continually learn the remaining 200 ImageNet classes in 4 rehearsal sessions (50 classes per rehearsal session). We set the number of iterations per rehearsal session to $1251$ for mini-batch size $512$ in latent rehearsal experiments. 
Whereas in veridical rehearsal experiments, we set the number of iterations per rehearsal session to $2502$ for mini-batch size $256$.
We implement Rainbow memory~\citep{bang2021rainbow}, MIR~\citep{aljundi2019online}, CLIB~\citep{koh2021online}, and ASER~\citep{shim2021online} following the corresponding papers and codes.

%\textbf{MobileNet/ImageNet-150 Experiments.}
\subsection{MobileNet/ImageNet-150 Experiments}
Here we specify settings used to compare gradient-based methods in Sec.~\ref{sec:grad_main}.
After base-initialization on 100 ImageNet classes, models continually learn the remaining 50 ImageNet classes in 5 rehearsal sessions (10 classes per rehearsal session). We set LR to $0.2$ and the number of iterations per rehearsal session to $500$ for mini-batch size $64$.
Other details adhere to the aforementioned ImageNet-1K experiments e.g., hyperparameters, OPQ, optimizer, LR scheduler, and pre-trained MobileNet architecture.
We implement OCS~\citep{yoononline} and GSS~\citep{aljundi2019gradient} following the corresponding papers and codes. We implement Grad matching~\citep{campbell2019automated} following the implementation from~\cite{borsos2020coresets}. 
%All methods use identical settings i.e., same pretrained model, same optimizer, same LR scheduler and same hyperparameters.

%\textbf{Representation Drift Experiments.}
\subsection{Representation Drift Experiments}
\label{sec:drift_detail}
We describe settings used in Sec.~\ref{sec:additional_exp} for the representation drift experiments. % where we analyze representation drift.
After base-initialization on 100 ImageNet classes, models continually learn 2 tasks each of which consists of 10 ImageNet classes. The number of iterations per task is $100$ for a mini-batch size of $512$.
Other details adhere to the aforementioned ImageNet-1K experiments e.g., hyperparameters, OPQ, optimizer, LR scheduler, and pre-trained MobileNet architecture.

Following~\citet{caccia2021new}, we measure the representation drift of an old sample $X$ at each training iteration $t$ as $\left\| f_{\theta_{t}}(X) - f_{\theta_{t+1}}(X) \right\|$ where $f_{\theta}$ denotes model parameters excluding the final layer. For this, we use a validation set of unseen old samples and their penultimate embedding vectors.
To compute Area Under the Curve (AUC), we use Scikit-learn's \textit{sklearn.metrics.auc} function.

%\textbf{Text Classification Experiments.}
\subsection{Text Classification Experiments.}
Here we describe settings used in continual text classification experiments in Sec.~\ref{sec:additional_exp}.
We use AdamW optimizer with LR of $3\times 10^{-5}$ and weight decay of $0.01$. We use batch size 8 and a maximum sequence length of 256. Other settings follow the replay baseline from IDBR paper~\citep{huang2021continual}.
We study a total of 6 task sequences.
They are: order 1 (ag $\rightarrow$ yelp $\rightarrow$ yahoo), order 2 (yelp $\rightarrow$ yahoo $\rightarrow$ ag), order 3 (yahoo $\rightarrow$ ag $\rightarrow$ yelp), order 4 (ag $\rightarrow$ yelp $\rightarrow$ amazon $\rightarrow$ yahoo $\rightarrow$ dbpedia), order 5 (yelp $\rightarrow$ yahoo $\rightarrow$ amazon $\rightarrow$ dbpedia $\rightarrow$ ag), and order 6 (dbpedia $\rightarrow$ yahoo $\rightarrow$ ag $\rightarrow$ amazon $\rightarrow$ yelp).

%\textbf{MobileNet/Places-LT Experiments.}
\subsection{MobileNet/Places-LT-365 Experiments}
These implementation details correspond to experiments in Sec.~\ref{sec:lt_exp}.
Since Places-LT is a small dataset, for base initialization, we adopt MobileNet backbone pre-trained on 100 ImageNet classes from MobileNet/ImageNet-1K experiment. 
We also adopt OPQ model pre-trained on same 100 ImageNet classes from MobileNet/ImageNet-1K experiment.
After base initialization, CL phase begins where models learn $365$ Places-LT classes in 5 rehearsal sessions ($73$ classes per rehearsal session) using SIESTA and latent rehearsal.
We use SGD optimizer and OneCycle LR scheduler with initial LR of $0.1$ and mini-batch size $32$. We set the number of iterations per rehearsal session to $1200$. 
Under memory constraints, memory is bounded by 20K samples. In an unconstrained memory setting, the entire dataset (62500 samples) is stored in a memory buffer.
Other settings follow MobileNet/ImageNet-1K experiments.
During the evaluation, we only use Places-LT-365 test set and do not use the 100 ImageNet classes subset used for base initialization.

%\textbf{MobileViT/ImageNet-300 Experiments.}
\subsection{MobileViT/ImageNet-300 Experiments.}
Here we describe the settings used in Sec.~\ref{sec:vit}.
Following SIESTA~\citep{harun2023siesta}, we use cosine cross entropy loss and replace batch norm with group norm and weight standardization in MobileViT-S architecture.
For universal feature extraction, we freeze the first 8 blocks including stem, 6 MobielNetV2 blocks, and 1 MobileViT block. We keep the remaining blocks (1 MobileNetV2 block and 2 MobileViT blocks) and layers (1 CNN layer and 1 linear layer) plastic during the continual learning phase.
Product quantization (OPQ) settings follow MobileNet/ ImageNet-1K experiments.

We use AdamW optimizer with initial LR of $4\times 10^{-4}$ and weight decay of $0.01$. We use OneCycle LR scheduler. During base initialization, we train MobileViT on 100 ImageNet classes using supervised pre-training for 300 epochs. For this, we use the same settings described above.
After base-initialization, models continually learn the remaining 200 ImageNet classes in 4 rehearsal sessions (50 classes per rehearsal session). We set the number of iterations per rehearsal session to 10K for mini-batch size 64.
Under memory constraints, memory is bounded by 130K samples. In an unconstrained memory setting, all 383708 samples are stored in a memory buffer.
MobileViT experiments are based on SIESTA and latent rehearsal.

\textbf{Compute.}
For compute (GPU) reasons, we vary the mini-batch size and number of iterations accordingly but compute constraints (iterations $\times$ mini-batch size) remain constant across experiments. We use a single GPU (NVIDIA RTX A5000) for all experiments.

\section{Dataset Details}
\label{sec:dataset}

We conduct vision experiments on ImageNet-1K and Places-LT.
\textbf{ImageNet-1K}~\citep{russakovsky2015imagenet} has 1000 categories, each with $732-1300$ training images and $50$ for validation. In total, it contains $1.28$ million training images and $50000$ test images.
\textbf{Places-LT}~\citep{liu2019large} is a long-tailed version of Places-2~\citep{zhou2017places}. Places-LT has $62500$ training images spanning $365$ classes with $5$ to $4980$ images each. For evaluation, we use the Places-LT validation set, which has 20 images per category ($7300$ total).
%\textbf{CUB-200}~\citep{welinder2010caltech} has RGB images of $200$ bird species with $5994$ training images and $5794$ test images. 
For NLP dataset details, we refer readers to IDBR paper~\citep{huang2021continual}.

\section{ImageNet-1K Curves for Old and New Tasks}
\begin{figure*}[h]
  \centering

\begin{subfigure}[b]{0.42\textwidth}
         \centering
         \includegraphics[width=\textwidth]{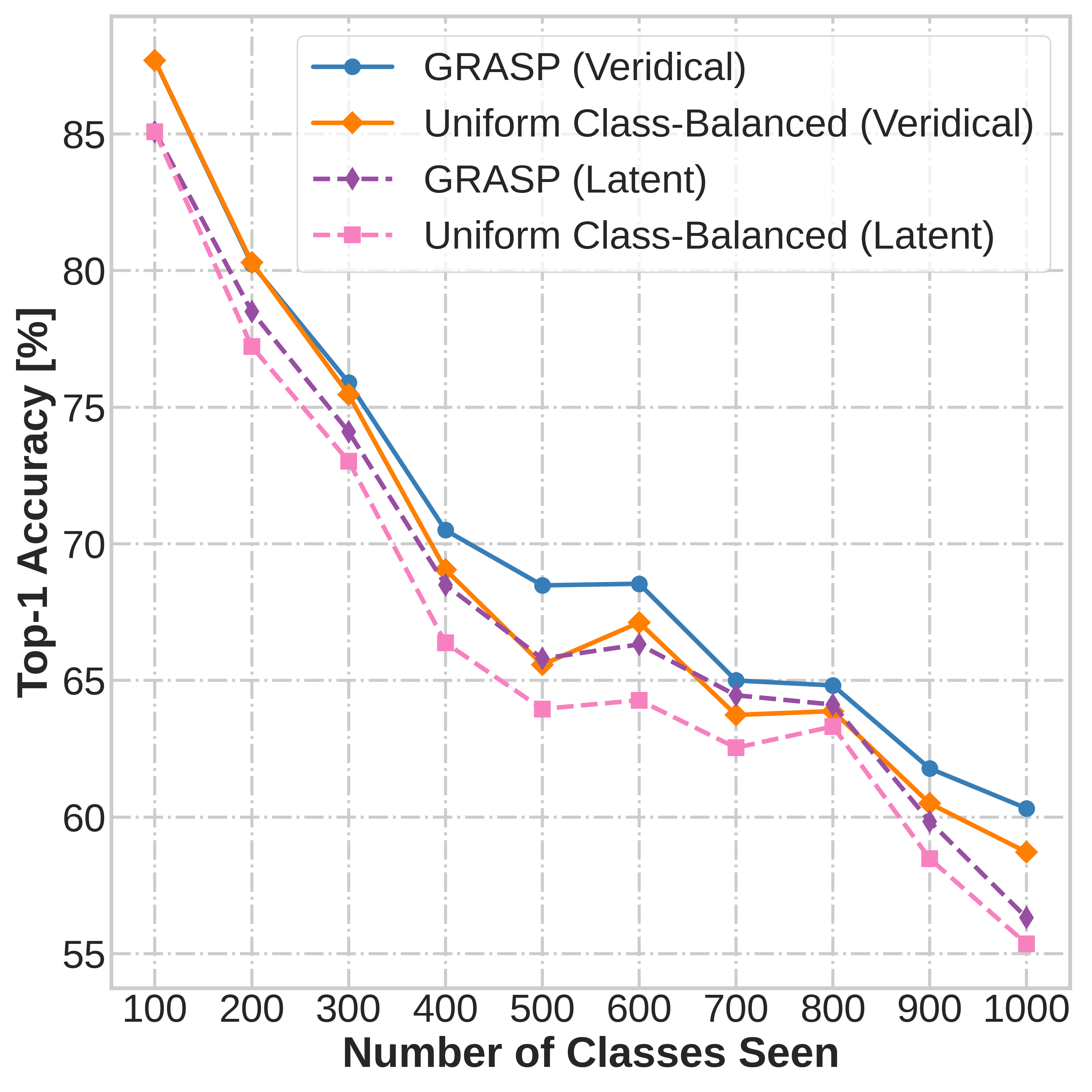}
         \caption{Accuracy on new classes (plasticity)}
         \label{fig:new}
     \end{subfigure}
     \hfill
      \begin{subfigure}[b]{0.42\textwidth}
         \centering
         \includegraphics[width=\textwidth]{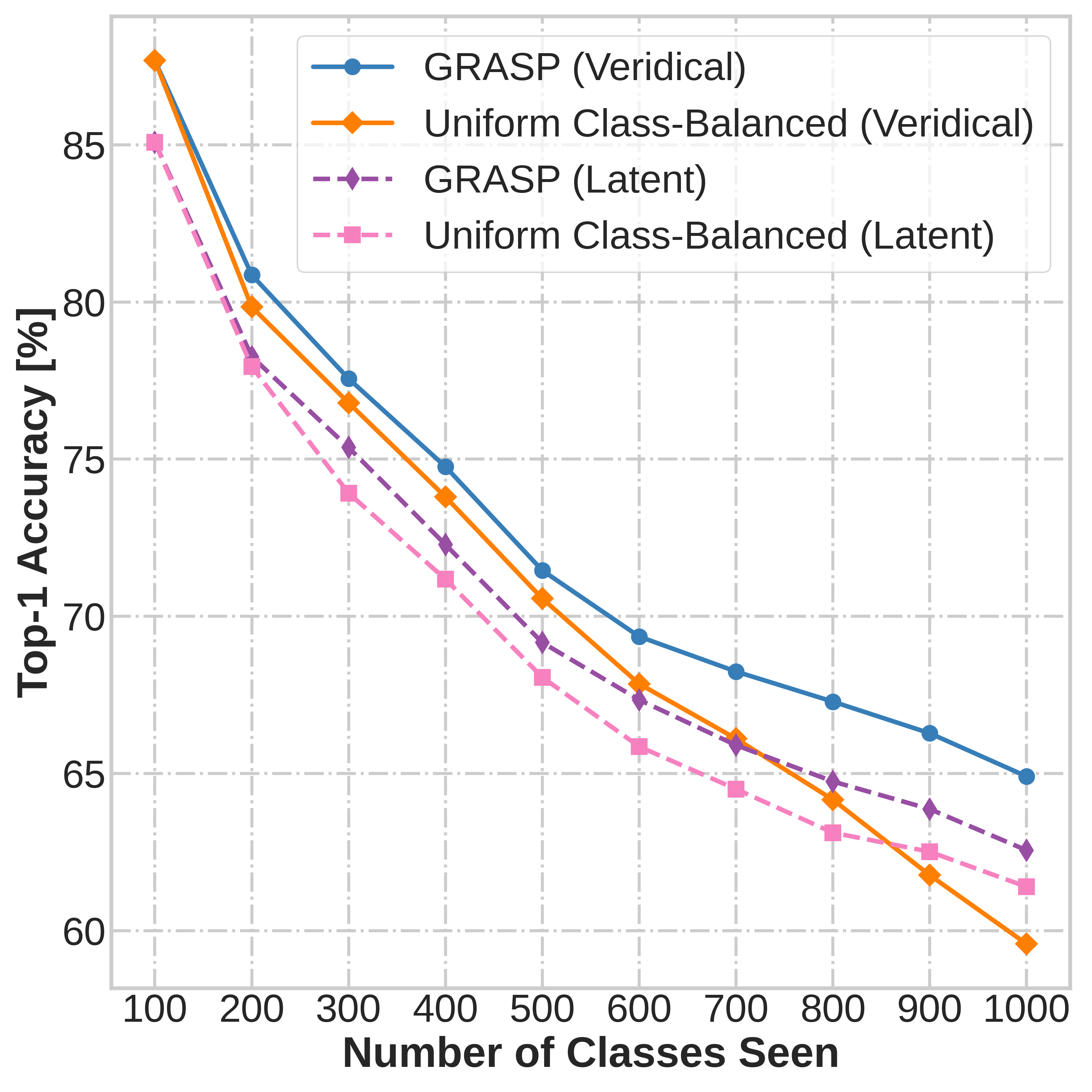}
         \caption{Accuracy on old classes (stability)}
         \label{fig:old}
     \end{subfigure}
     \hfill
    % \begin{subfigure}[b]{0.3\textwidth}
    %     \centering
    %     \includegraphics[width=\textwidth]{images/comp_acc_all.png}
    %     \caption{Accuracy on all classes}
    %     \label{fig:all}
    % \end{subfigure}
   \caption{Performance of latent rehearsal policies on old and new classes in online CIL on ImageNet-1K. All methods use SIESTA and the same pre-trained MobileNet architecture.
   \label{fig:old_new}}
   %\vspace{-2em}
\end{figure*}

In the main text, we showed ImageNet-1K learning curves for all seen tasks (Fig.~\ref{fig:in1k}).
Here we show ImageNet-1K curves for old and new tasks in Fig.~\ref{fig:old_new}. These experiments use our default MobileNet/ImageNet-1K setup with SIESTA.
We find that GRASP shows better performance than uniform balanced in all rehearsal sessions (100 ImageNet classes per rehearsal session) for both old and new tasks. This demonstrates that GRASP maintains a good balance between stability (old task) and plasticity (new task).

 % Fig showing performance on old and new tasks

\begin{table}[h]
  \caption{\textbf{Long-Tailed Recognition (Places-LT-365)}. Comparison between GRASP and uniform balanced in CIL on Places-LT with SIESTA and latent rehearsal.
  Here $\mu_{N}$, $\mu_{O}$, and $\mu_{A}$ denote accuracy ($\%$) on new, old, and all classes respectively averaged over rehearsal sessions. And, $\alpha_{H}$, $\alpha_{T}$, and $\alpha$ stand for final accuracy ($\%$) on head ($>100$ examples), tail ($\leq 100$ examples), and all classes respectively.
  Reported results are averaged over 6 runs. %The $(\uparrow)$ indicates high value is desirable. 
  Uniform balanced rehearsal is referred as Uniform$\dag$. 
  }
  \centering
  \resizebox{\linewidth}{!}{
     \begin{tabular}{l|cccccc|cccccc}
     \hline
     \multicolumn{1}{c|}{\textbf{Method}} &
     \multicolumn{6}{c|}{\textbf{Unbounded Memory}} &
     \multicolumn{6}{c}{\textbf{Bounded Memory}} \\
      & $\mu_{N} \uparrow$ & $\mu_{O} \uparrow$ & $\mu_{A} \uparrow$ & $\alpha_{H} \uparrow$ & $\alpha_{T} \uparrow$ & $\alpha \uparrow$ & $\mu_{N} \uparrow$ & $\mu_{O} \uparrow$ & $\mu_{A} \uparrow$ & $\alpha_{H} \uparrow$ & $\alpha_{T} \uparrow$ & $\alpha \uparrow$ \\
     %\hline
     %Upper Bound & --- & --- & --- & --- & --- & $22.49$ & --- & --- & --- & --- & --- & $22.49$ \\
     \hline
     Uniform$\dag$ & $35.09$ & $33.19$ & $33.72$ & $13.41$ & $14.53$ & $25.70$ & $35.13$ & $32.63$ & $33.37$ & $12.71$ & $14.44$ & $24.91$ \\
     \textbf{GRASP} & $\mathbf{35.51}$ & $\mathbf{33.37}$ & $\mathbf{34.02}$ & $\mathbf{13.63}$ & $\mathbf{14.67}$ & $\mathbf{25.99}$ & $\mathbf{35.48}$ & $\mathbf{33.41}$ & $\mathbf{34.03}$ & $\mathbf{12.83}$ & $\mathbf{14.86}$ &  $\mathbf{25.54}$ \\
     \hline
     %Uniform$\dag$ & $35.13$ & $32.63$ & $33.37$ & $12.71$ & $14.44$ & $24.91$ \\
     %\textbf{GRASP} & $\mathbf{35.48}$ & $\mathbf{33.41}$ & $\mathbf{34.03}$ & $\mathbf{12.83}$ & $\mathbf{14.86}$ &  $\mathbf{25.54}$ \\
     %\hline
    \end{tabular}}
  \label{tab:places-LT}
\end{table}

\section{Places-LT-365 Results}
\label{sec:lt_exp}
Here, we use our default MobileNet/Places-LT setup with SIESTA and latent rehearsal.
We run each compared method 6 times using 6 data orderings and report the average results in Table~\ref{tab:places-LT}.
In long-tailed recognition, GRASP consistently outperforms uniform balanced in all evaluation criteria for both unbounded and bounded memory settings. Therefore GRASP demonstrates robustness to long-tailed data distributions.

\section{MobileViT Results}
\label{sec:vit}
For this analysis, we use our default MobileViT/ImageNet-300 setup with SIESTA and latent rehearsal.
Previously, we evaluated GRASP using CNN, now we evaluate GRASP using ViT. Table~\ref{tab:vit} summarizes the results. 
We observe that GRASP achieves higher accuracy than uniform balanced in all metrics for both unbounded and bounded memory settings. This indicates that GRASP generalizes to ViT architecture besides CNN.

\begin{table}[t]
  \caption{ 
  \textbf{MobileViT Results (ImageNet-300).} Comparison between GRASP and uniform balanced in CIL on ImageNet-300 with SIESTA and latent rehearsal.
   Here $\mu_{N}$, $\mu_{O}$, and $\mu_{A}$ denote accuracy ($\%$) on new, old, and all classes respectively averaged over rehearsal sessions. And $\alpha$ is the final accuracy ($\%$) on all classes.}
  \centering
     \begin{tabular}{l|cccc|cccc}
     \hline
     \multicolumn{1}{c|}{\textbf{Method}} &
     \multicolumn{4}{c|}{\textbf{Unbounded Memory}} &
     \multicolumn{4}{c}{\textbf{Bounded Memory}} \\
      & $\mu_{N} \uparrow$ & $\mu_{O} \uparrow$ & $\mu_{A} \uparrow$ & $\alpha \uparrow$ & $\mu_{N} \uparrow$ & $\mu_{O} \uparrow$ & $\mu_{A} \uparrow$ & $\alpha \uparrow$ \\
     \hline
     Uniform Bal & $67.60$ & $71.43$ & $70.53$ & $64.19$ & $65.82$ & $68.47$ & $67.75$ & $59.01$ \\
     \textbf{GRASP} & $\mathbf{68.18}$ & $\mathbf{72.05}$ & $\mathbf{71.15}$ & $\mathbf{65.37}$ & $\mathbf{66.54}$ & $\mathbf{68.86}$ & $\mathbf{68.20}$ & $\mathbf{59.26}$ \\
     \hline
    \end{tabular}
  \label{tab:vit}
\end{table} % MobileViT experiments

\begin{table}[h]
  %\footnotesize
  \caption{ 
  \textbf{Continual Text Classification (3 runs).}
   GRASP versus the uniform balanced in continual text classification. %An offline trained multi-task learning model serves as an upper bound. 
   %Both compute and memory constraints are imposed \ck{can delete this sentence if you say it in the text}. 
   %Performance is averaged over 3 runs.
   }
  % \label{tab:comp_nlp}
  \centering
  \resizebox{\linewidth}{!}{
     \begin{tabular}{c|cccc|cccc}
     %\begin{tabular}{l|llll|llll}
     \hline
     \multicolumn{1}{c|}{\textbf{Method}} &
     \multicolumn{4}{c|}{\textbf{Length-3 Task Sequences}} &
     \multicolumn{4}{c}{\textbf{Length-5 Task Sequences}} \\
     \hline
     \textbf{Order} & \textbf{1} & \textbf{2} & \textbf{3} & \textbf{Avg.} & \textbf{4} & \textbf{5} & \textbf{6} & \textbf{Avg.} \\
     \hline
     %Upper Bound & $74.16$ & $74.16$ & $74.16$ & $74.16$ & $75.09$ & $75.09$ & $75.09$ & $75.09$ \\
     %\hline
     % seed 0 \\ seed 1 \\ seed 3
     %\hline
     % buffer 50%
     %\hline
     %Uniform & $72.73$ & $73.32$ & $72.57$ & -- & $74.39$ & $73.75$ & $73.96$ & -- \\
     %GRASP & $73.65$ & $72.88$ & $73.22$ & -- & $74.95$ & $74.17$ & $73.97$ & -- \\
     %Uniform & $72.75$ & $73.52$ & $72.61$ & -- & $74.32$ & $74.71$ & $74.00$ & -- \\
     %GRASP & $73.70$ & $72.95$ & $73.71$ & -- & $74.58$ & $74.30$ & $75.05$ & -- \\
     %Uniform & $73.70$ & $73.16$ & $72.97$ & -- & $74.28$ & $73.98$ & $74.50$ & -- \\
     %GRASP & $73.42$ & $73.05$ & $72.84$ & -- & $74.47$ & $74.20$ & $74.07$ & -- \\
     %\hline
     Uniform Bal & $73.06$ \tiny $\pm 0.45$ & $\mathbf{73.33}$ \tiny $\pm 0.15$ & $72.72$ \tiny $\pm 0.18$ & $73.04$ & $74.33$ \tiny $\pm 0.05$ & $74.15$ \tiny $\pm 0.41$ & $74.15$ \tiny $\pm 0.25$ & $74.21$ \\
     \textbf{GRASP} & $\mathbf{73.59}$ \tiny $\pm 0.12$ & $72.96$ \tiny $\pm 0.07$ & $\mathbf{73.26}$ \tiny $\pm 0.36$ & $\mathbf{73.27}$ & $\mathbf{74.67}$ \tiny $\pm 0.21$ & $\mathbf{74.22}$ \tiny $\pm 0.06$ & $\mathbf{74.36}$ \tiny $\pm 0.49$ & $\mathbf{74.42}$ \\
     \hline
    \end{tabular}}
  %\caption{ 
  %\textbf{Continual Text Classification.}
  % GRASP versus the uniform rehearsal policy in continual text classification. An offline trained multi-task learning model serves as an upper bound. 
   %Both compute and memory constraints are imposed \ck{can delete this sentence if you say it in the text}. 
  % Performance is averaged over 3 runs.}
    \label{tab:comp_nlp}
\end{table} % Text classification results

\section{Continual Text Classification Results}
\label{sec.text_classification}

Due to space limitations in the main paper, we include the continual text classification results in this section. As shown in Table~\ref{tab:comp_nlp}, GRASP surpasses uniform balanced rehearsal in 5 out of 6 task sequences.

%\section{Representation Drift Quantitative Results}
%\label{sec.drift_results}

%\input{tables/comp_drift}

%Due to space limitations in the main paper, we include the representation drift results in this section. As shown in Table~\ref{tab:comp_dirft}, GRASP achieves the lowest drift among compared methods. Implementation details are given in Appendix~\ref{sec:drift_detail}.

\section{Qualitative Analysis}
\label{sec.vis}

\begin{figure}[t]
    \centering
    \begin{subfigure}[b]{0.42\textwidth}
        \centering
        \includegraphics[width=\textwidth]{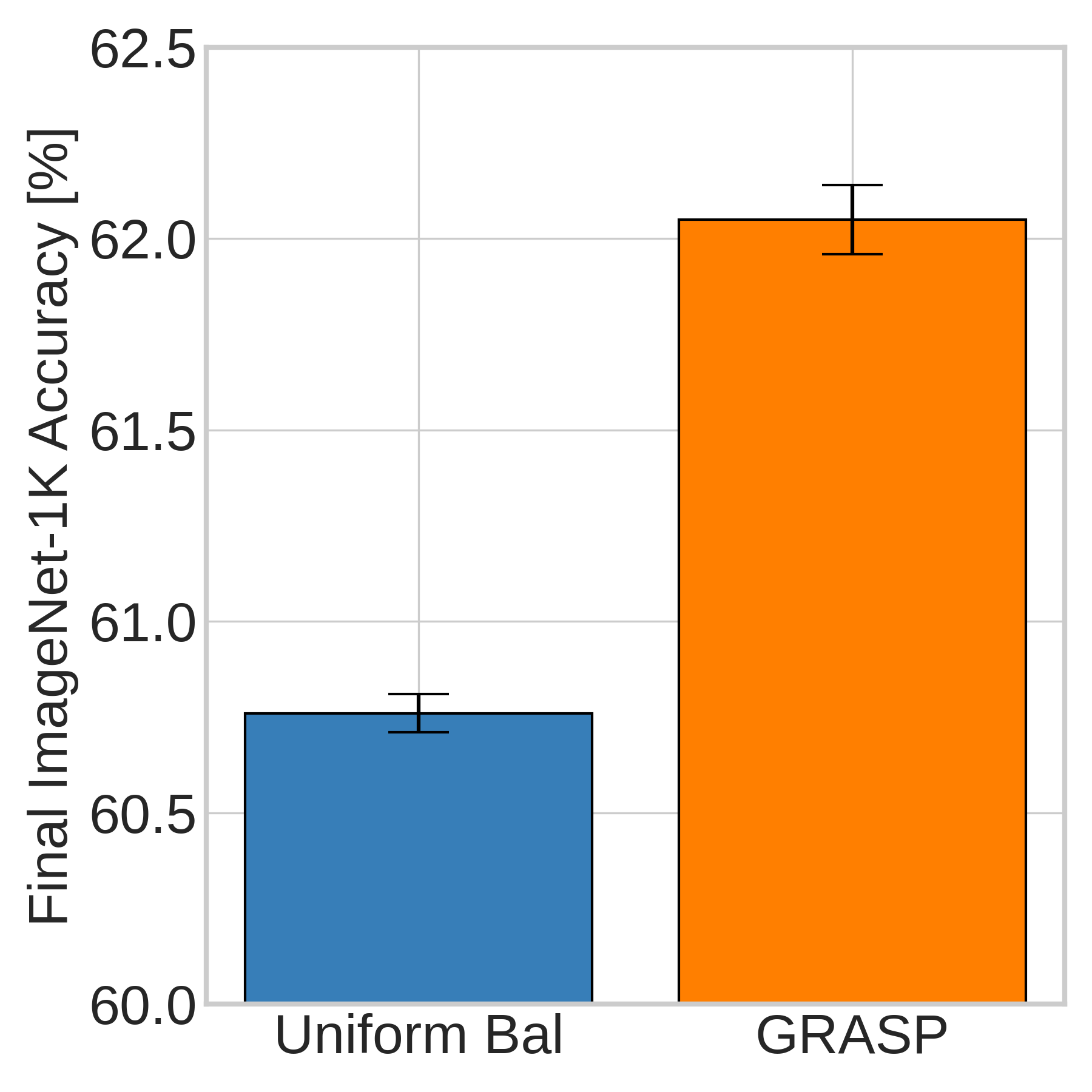}
        \caption{\textbf{Latent Rehearsal (Unbounded Memory)}}
        \label{fig:latent_unbounded}
    \end{subfigure}
    \hfill 
    \begin{subfigure}[b]{0.42\textwidth}
        \centering
        \includegraphics[width=\textwidth]{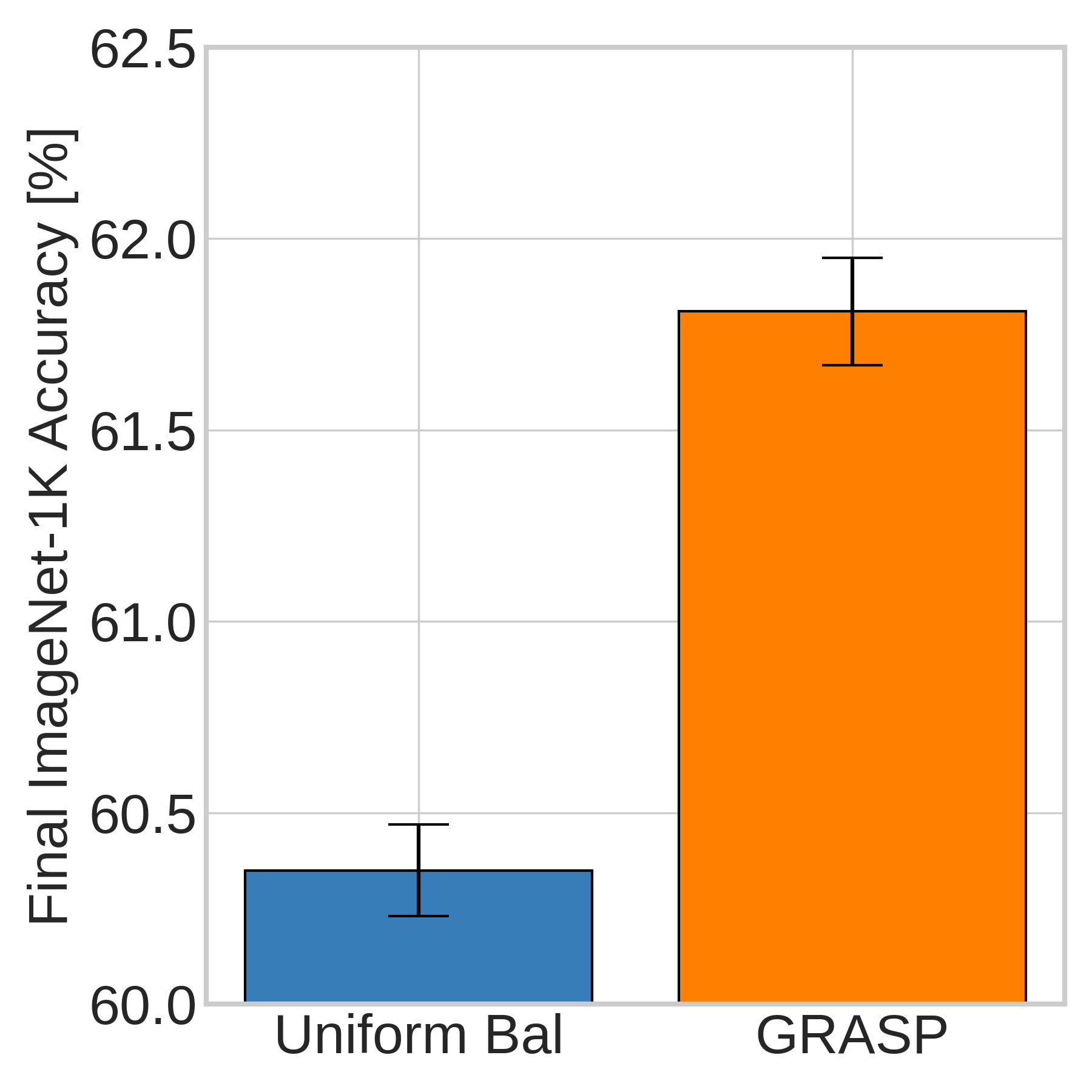}
        \caption{\textbf{Latent Rehearsal (Bounded Memory)}}
        \label{fig:latent_bounded}
    \end{subfigure}

    % Add a little vertical space between rows
    \vspace{1em}

    \begin{subfigure}[b]{0.42\textwidth}
        \centering
        \includegraphics[width=\textwidth]{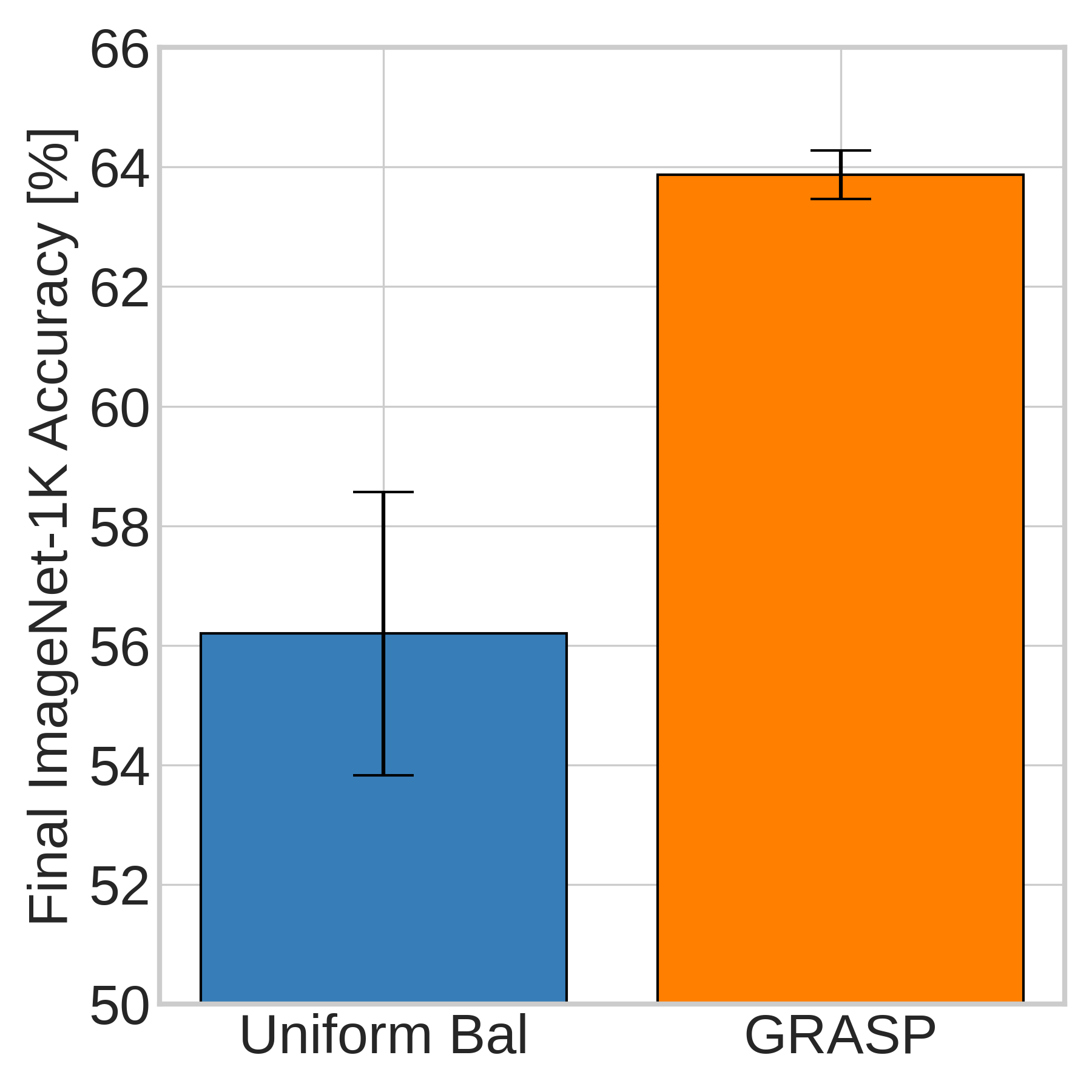}
        \caption{\textbf{Veridical Rehearsal (Unbounded Memory)}}
        \label{fig:veridical_unbounded}
    \end{subfigure}
    \hfill 
    \begin{subfigure}[b]{0.42\textwidth}
        \centering
        \includegraphics[width=\textwidth]{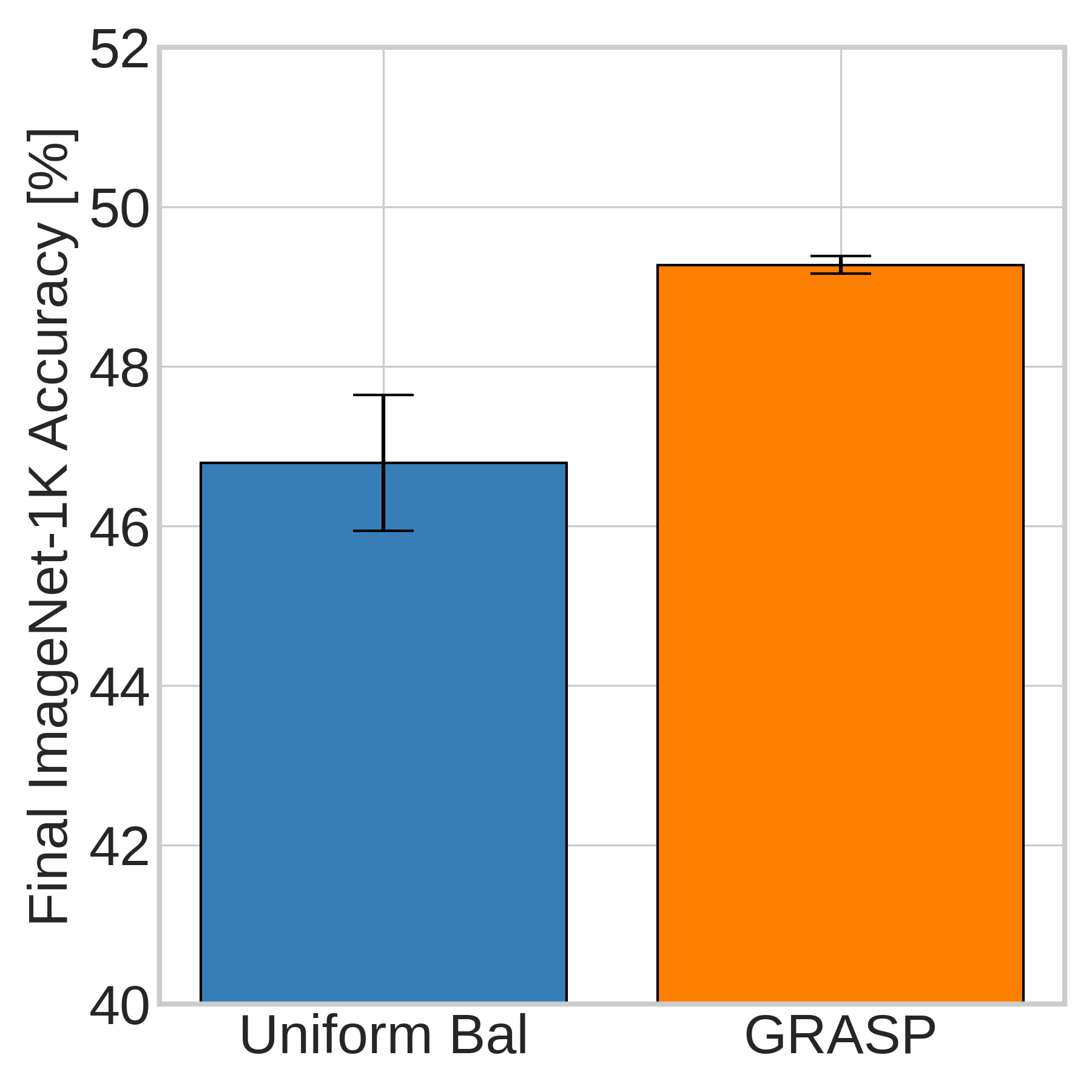}
        \caption{\textbf{Veridical Rehearsal (Bounded Memory)}}
        \label{fig:veridical_bounded}
    \end{subfigure}
        \caption{
        Qualitative comparison between GRASP and uniform balanced in CIL on ImageNet-1K with SIESTA. Each plot shows final accuracy (\%) on ImageNet-1K averaged over 3 runs while indicating standard deviation ($\pm$). 
        }
        \label{fig:barplots_comp}
\end{figure}

\begin{figure}[t]
    \centering
    \begin{subfigure}[b]{0.42\textwidth}
        \centering
        \includegraphics[width=\textwidth]{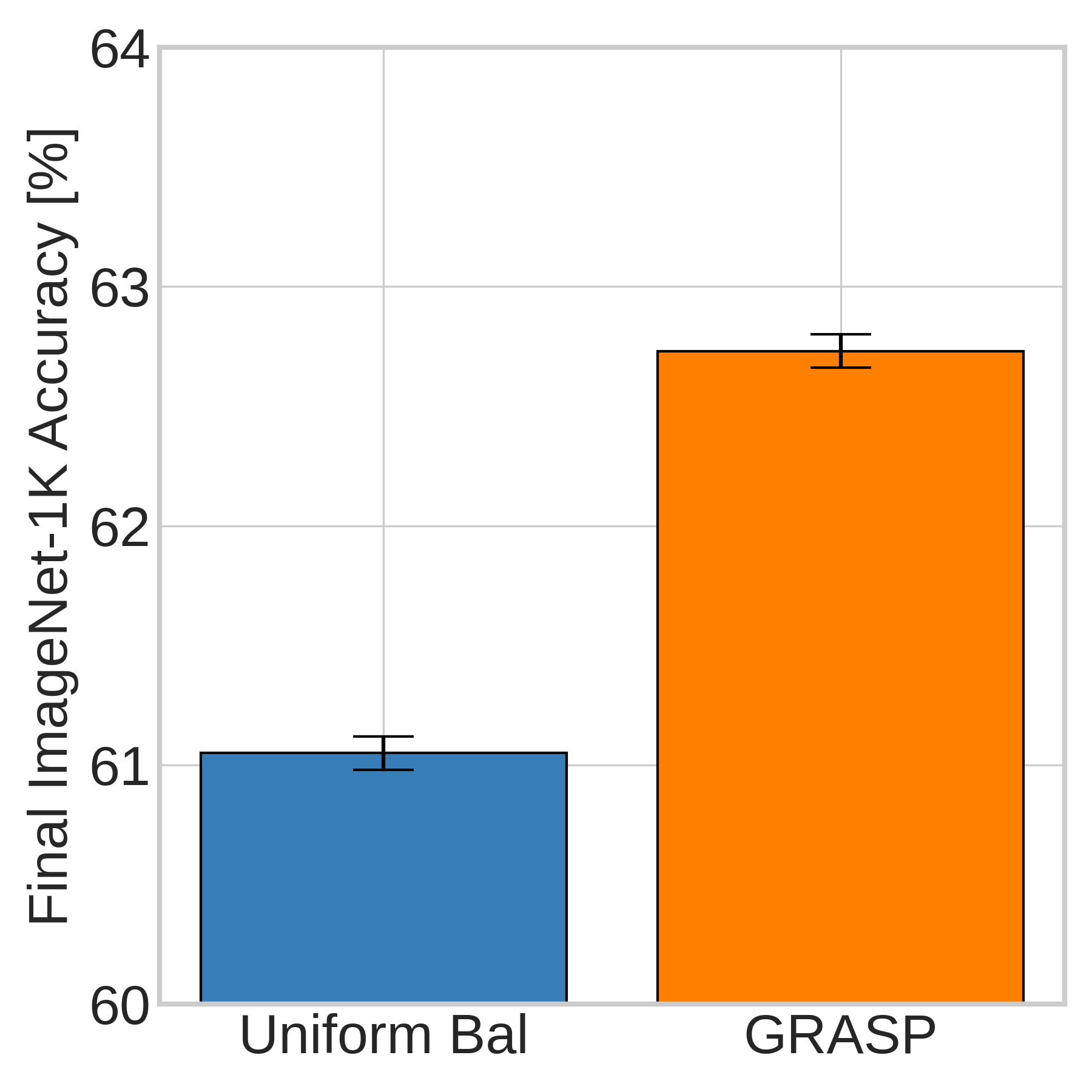}
        \caption{\textbf{Latent Rehearsal (GDumb)}}
        \label{fig:gdumb_latent_unbounded}
    \end{subfigure}
    \hfill 
    \begin{subfigure}[b]{0.42\textwidth}
        \centering
        \includegraphics[width=\textwidth]{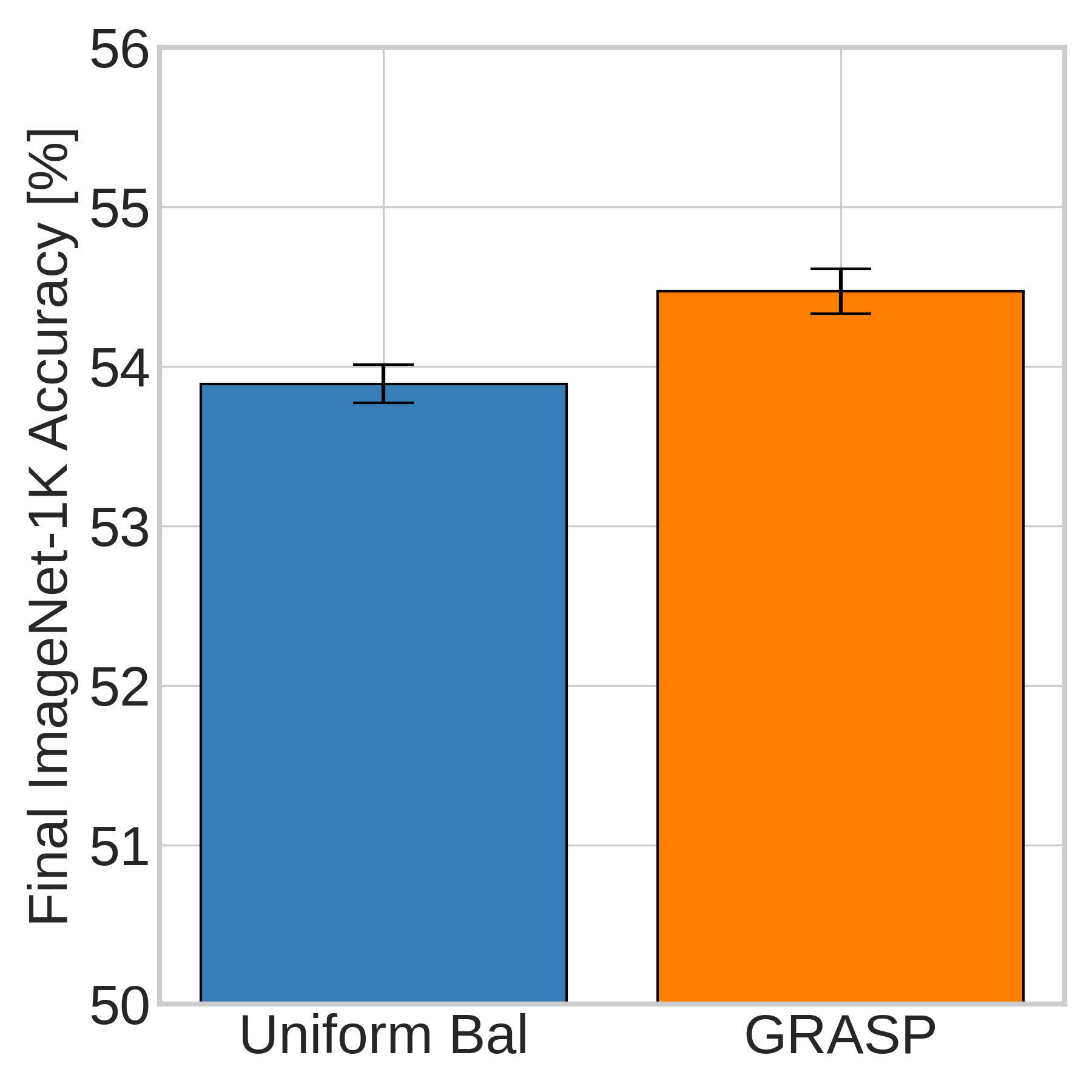}
        \caption{\textbf{Latent Rehearsal (DERpp)}}
        \label{fig:derpp_latent_bounded}
    \end{subfigure}

    % Add a little vertical space between rows
    \vspace{1em}

    \begin{subfigure}[b]{0.42\textwidth}
        \centering
        \includegraphics[width=\textwidth]{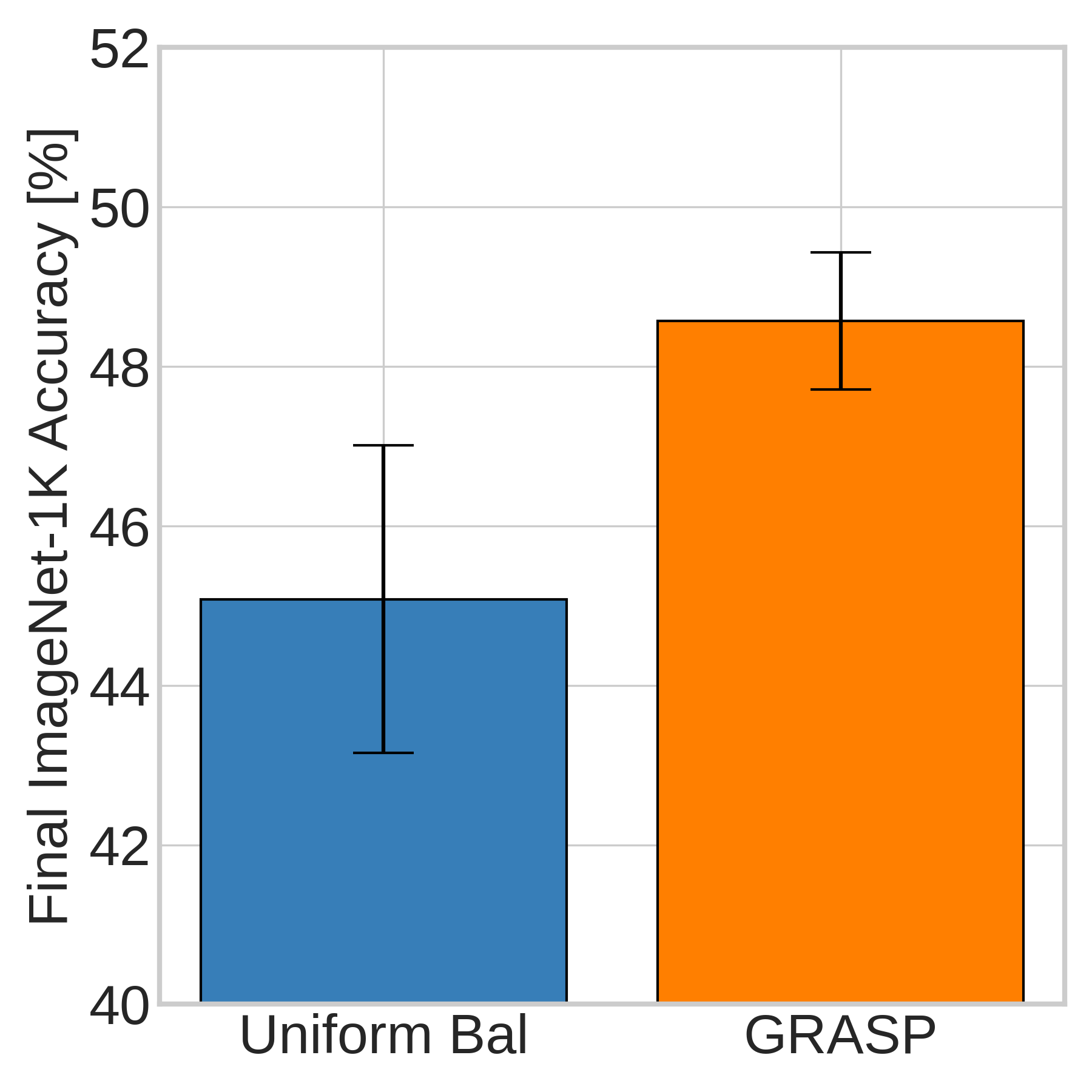}
        \caption{\textbf{Veridical Rehearsal (GDumb)}}
        \label{fig:derpp_veridical_unbounded}
    \end{subfigure}
    \hfill 
    \begin{subfigure}[b]{0.42\textwidth}
        \centering
        \includegraphics[width=\textwidth]{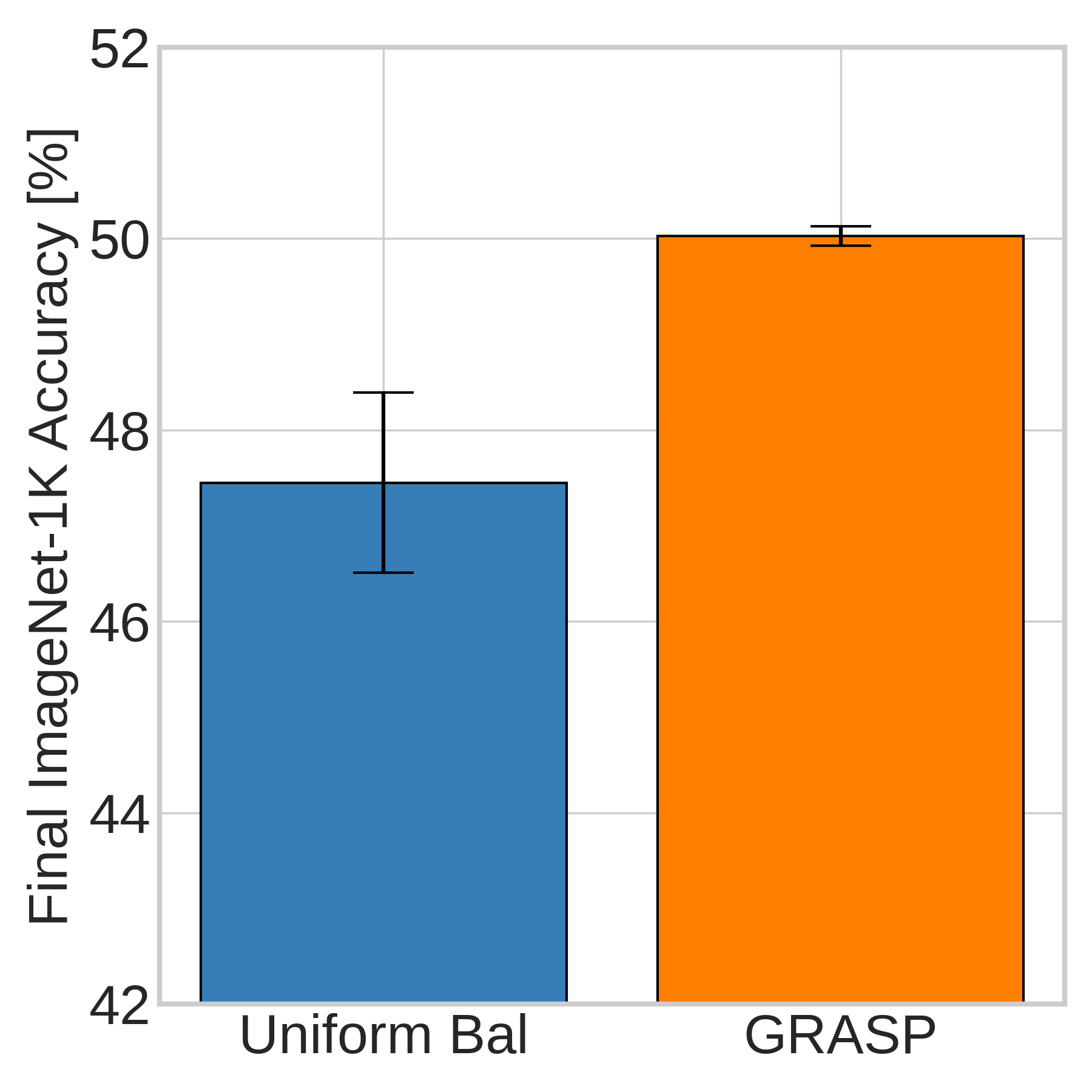}
        \caption{\textbf{Veridical Rehearsal (DERpp)}}
        \label{fig:gdumb_veridical_bounded}
    \end{subfigure}
        \caption{
        Qualitative comparison between GRASP and uniform balanced in memory-bounded CIL on ImageNet-1K with GDumb and DERpp. Each plot shows final accuracy (\%) on ImageNet-1K averaged over 3 runs while indicating standard deviation ($\pm$). 
        }
        \label{fig:barplots_comp2}
\end{figure}

\begin{figure}[t]
  \centering
   \includegraphics[width=0.45\linewidth]{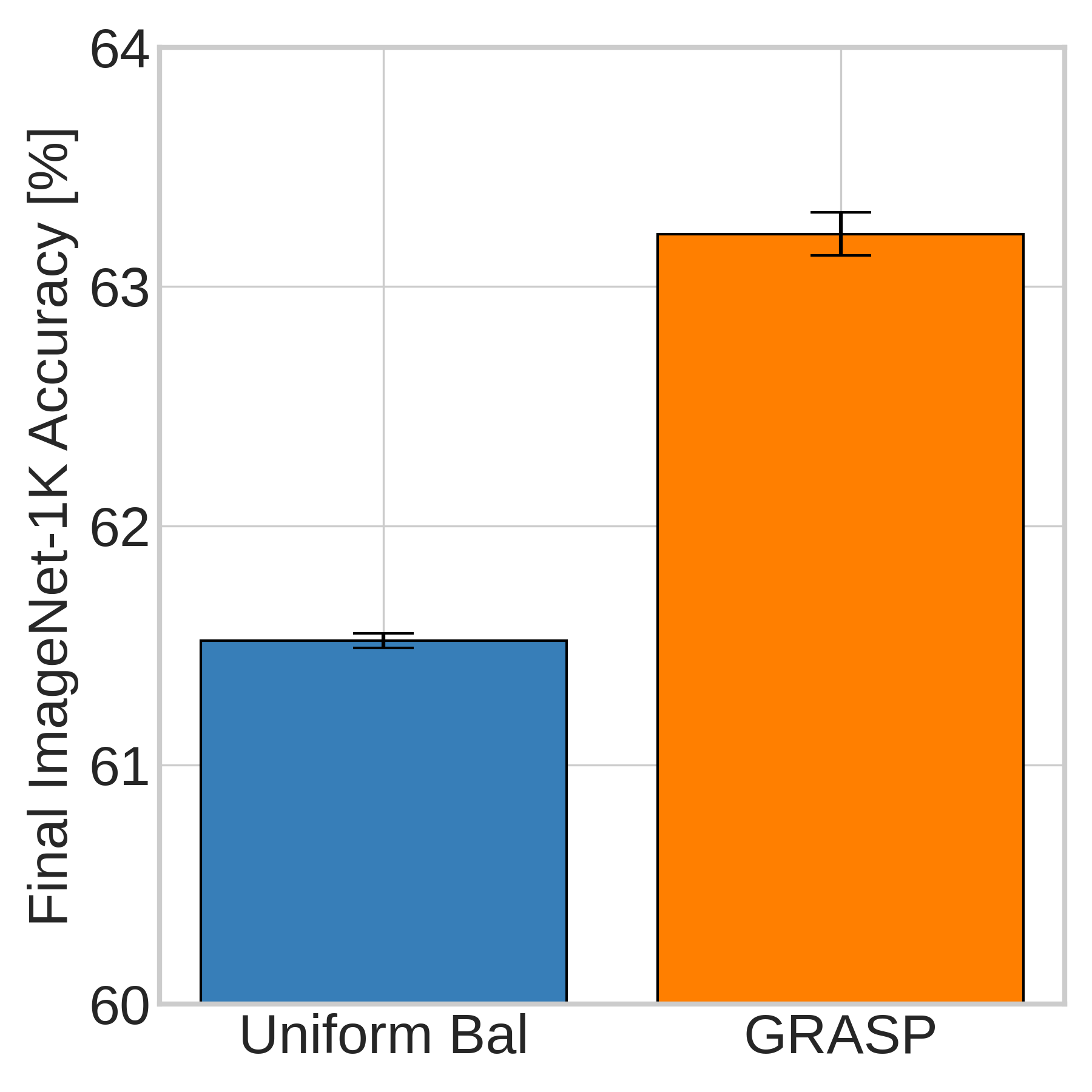}
   \caption{
        Qualitative comparison between GRASP and uniform balanced in memory-bounded IID CL experiments on ImageNet-1K using SIESTA with latent rehearsal. Each plot shows final accuracy (\%) on ImageNet-1K averaged over 3 runs while indicating standard deviation ($\pm$). 
        }
   \label{fig:iid_vis}
\end{figure}

In Sec.~\ref{sec:imagenet1k_main}, we summarized all the ImageNet-1K results where GRASP outperformed uniform balanced rehearsal policy. In this section, we present bar plots to analyze the performance improvements of GRASP over uniform balanced rehearsal policy in various ImageNet-1K experiments. As shown in Fig.~\ref{fig:barplots_comp}, Fig.~\ref{fig:barplots_comp2}, and Fig.~\ref{fig:iid_vis}, GRASP outperforms uniform balanced by nontrivial margins in all ImageNet-1K experiments. 

\begin{figure}[t]
  \centering
   \includegraphics[width=0.9\linewidth]{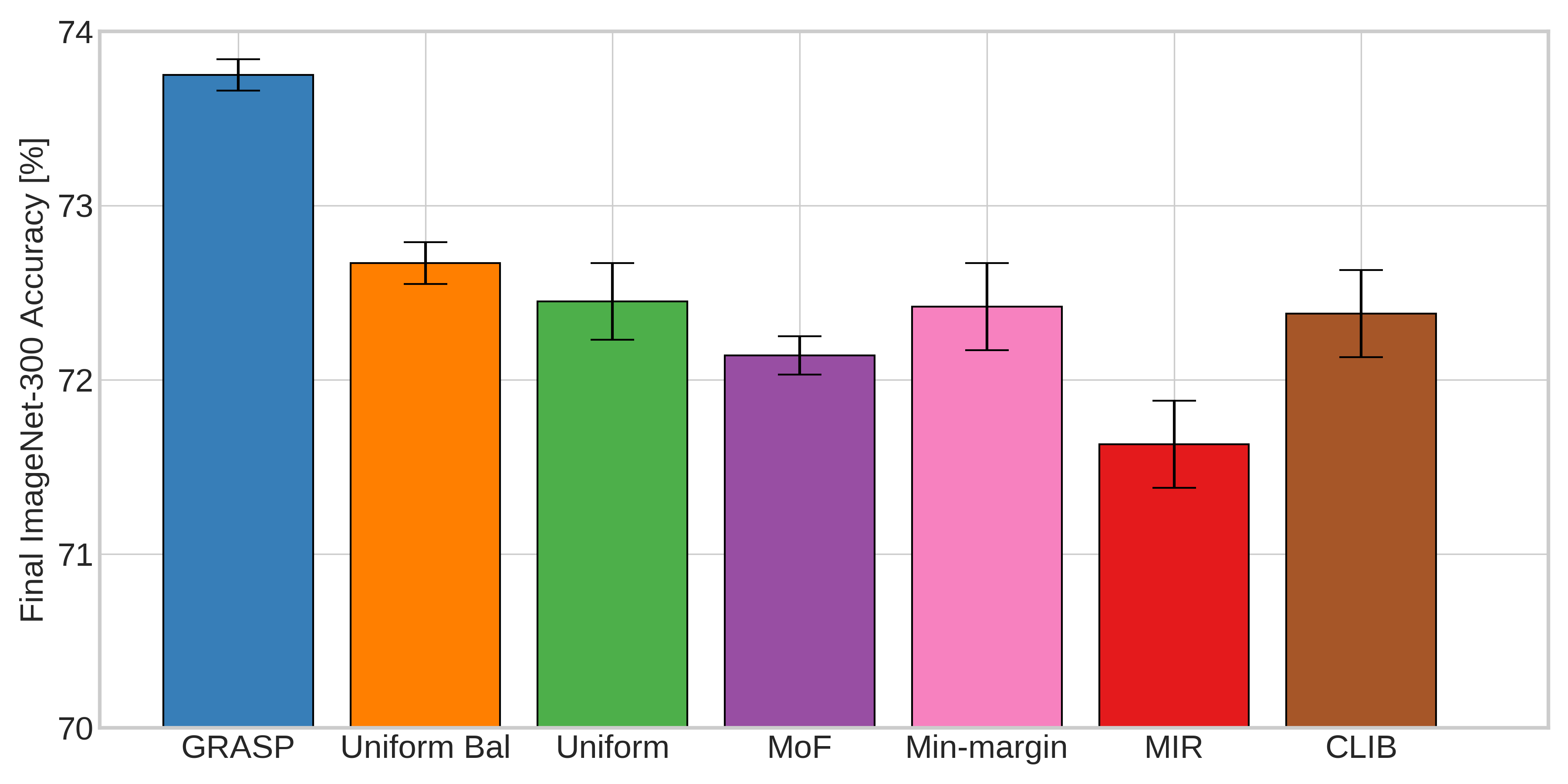}
   \caption{
        Qualitative comparison between GRASP and other competitive policies in memory-unbounded CIL experiments on ImageNet-300 using SIESTA with latent rehearsal. Each plot shows final accuracy (\%) on ImageNet-300 averaged over 3 runs while indicating standard deviation ($\pm$). 
        }
   \label{fig:cil_imagenet300_vis}
\end{figure}

In Sec.~\ref{sec:imagenet_subset_main}, we summarized all the ImageNet-300 results where GRASP outperforms various rehearsal policies. Here, we qualitatively compare GRASP with the performant policies. As illustrated in Fig.~\ref{fig:cil_imagenet300_vis}, GRASP outperforms other competitive policies in CIL experiments on ImageNet-300.

\end{document}